%% file: main.tex
\pdfoutput=1
\documentclass[fleqn,10pt]{wlscirep}

\usepackage[utf8]{inputenc}
\usepackage[T1]{fontenc}

\usepackage{algorithmic}
\usepackage{algorithm}
\usepackage{amsfonts}
\usepackage{amsmath}
\usepackage{amssymb}
\usepackage{amsthm}
\usepackage{array}
\usepackage{arydshln}
\usepackage{bm}
\usepackage{cite}
\usepackage{color}
\usepackage{comment}
\usepackage{dsfont}
\usepackage{enumitem}
\usepackage{float}
\usepackage[T1]{fontenc}
\usepackage{geometry}
\usepackage{graphicx}
\usepackage{grffile}
\usepackage{hyperref}\hypersetup{colorlinks,linkcolor=blue,anchorcolor=blue,citecolor=blue}
\usepackage{letltxmacro}
\usepackage{mathrsfs}
\usepackage{mathtools}
\usepackage{multirow}
\usepackage{nicefrac}
\usepackage{subcaption}
\usepackage{tablefootnote}
\usepackage{verbatim}
\usepackage{booktabs}

\usepackage{todonotes}
\newcommand{\bA}{\bm{A}}

\newcommand{\bK}{\bm{K}}

\newcommand{\bQ}{\bm{Q}}

\newcommand{\bV}{\bm{V}}
\newcommand{\bW}{\bm{W}}
\newcommand{\bX}{\bm{X}}
\newcommand{\bY}{\bm{Y}}

\newcommand{\cL}{\mathcal{L}}

\newcommand{\cO}{\mathcal{O}}

\newcommand{\bcA}{\bm{\mathcal{A}}}
\newcommand{\bcB}{\bm{\mathcal{B}}}
\newcommand{\bcC}{\bm{\mathcal{C}}}

\newcommand{\bcE}{\bm{\mathcal{E}}}

\newcommand{\bcM}{\bm{\mathcal{M}}}

\newcommand{\bcX}{\bm{\mathcal{X}}}

\newcommand{\RR}{\mathbb{R}}

\DeclareMathOperator{\fro}{\mathsf{F}}

\allowdisplaybreaks
\makeatletter

\theoremstyle{plain} 

\theoremstyle{definition}
 
\theoremstyle{remark}

\definecolor{tian}{RGB}{0,150,0}
\definecolor{cm}{RGB}{250,0,200}
\definecolor{yc}{RGB}{255,0,0}
\definecolor{hd}{RGB}{0,180,200}

\title{A Lightweight Transformer for Faster and Robust EBSD Data Collection}

\author[1, *]{Harry Dong}
\author[2]{Sean Donegan}
\author[2]{Megna Shah}
\author[1]{Yuejie Chi}
\affil[1]{Carnegie Mellon University, Department of Electrical and Computer Engineering, Pittsburgh, 15289, USA}
\affil[2]{Air Force Research Laboratory, Materials and Manufacturing Directorate, Wright-Patterson AFB, 45433, USA}

\affil[*]{harryd@andrew.cmu.edu}

\input{abstract}

\begin{document}

\flushbottom
\maketitle
\thispagestyle{empty}
% \noindent Please note: Abbreviations should be introduced at the first mention in the main text – no abbreviations lists. Suggested structure of main text (not enforced) is provided below.

\input{introduction}

\input{background}

\input{method}

\input{experiments}

\input{conclusion}

\section*{Data Availability Statement}

The IN625 and IN718 datasets are available in the repositories, \url{https://acdc.alcf.anl.gov/mdf/detail/shade_afrl_am_package_v2.1/} and \url{https://datadryad.org/stash/dataset/doi:10.5061/dryad.83bk3j9sj}, respectively. Synthetic EBSD data are generated using DREAM.3D and are located at \url{https://drive.google.com/drive/u/1/folders/1S-jKZ7wxIT4ra4q5VJ_vEl4rO3yYQSgQ}.

\bibliography{refs}

% \section*{Results}

% Up to three levels of \textbf{subheading} are permitted. Subheadings should not be numbered.

% \subsection*{Subsection}

% Example text under a subsection. Bulleted lists may be used where appropriate, e.g.

% \begin{itemize}
% \item First item
% \item Second item
% \end{itemize}

% \subsubsection*{Third-level section}
 
% Topical subheadings are allowed.

% \section*{Discussion}

% The Discussion should be succinct and must not contain subheadings.

% \section*{Methods}

% Topical subheadings are allowed. Authors must ensure that their Methods section includes adequate experimental and characterization data necessary for others in the field to reproduce their work.

% \bibliography{sample}

% \noindent LaTeX formats citations and references automatically using the bibliography records in your .bib file, which you can edit via the project menu. Use the cite command for an inline citation, e.g.  \cite{Hao:gidmaps:2014}.

% For data citations of datasets uploaded to e.g. \emph{figshare}, please use the \verb|howpublished| option in the bib entry to specify the platform and the link, as in the \verb|Hao:gidmaps:2014| example in the sample bibliography file.

\section*{Acknowledgements}

This work is supported in part by the Air Force D3OM2S Center of Excellence under FA8650-19-2-5209, and by the Carnegie Mellon University Manufacturing Futures Initiative, made possible by the Richard King Mellon Foundation. The work of H. Dong is also supported by Liang Ji-Dian Graduate Fellowship and Michel and Kathy Doreau Graduate Fellowship in Electrical and Computer Engineering at Carnegie Mellon University. The U.S. Government is authorized to reproduce and distribute reprints for Governmental purposes notwithstanding any copyright notation thereon. Authors thank Michael Uchic, Marc DeGraef, Greg Rohrer, and Zachary Varley for engaging discussions throughout the course of this work. Diagrams were created in draw.io 21.5.1.

\section*{Author contributions statement}

% Must include all authors, identified by initials, for example:
% A.A. conceived the experiment(s),  A.A. and B.A. conducted the experiment(s), C.A. and D.A. analysed the results.  All authors reviewed the manuscript. 

H.D. designed and implemented the model, projection algorithm, and experiments. S.D. and M.S. processed the raw data and helped integrate domain knowledge to our method. S.D., M.S., and Y.C. advised and guided the project. All authors wrote and reviewed the manuscript. 

\section*{Additional information}

% To include, in this order: \textbf{Accession codes} (where applicable); 
\textbf{Competing interests:} 
The authors declare no competing interests.

\end{document}

%% file: abstract.tex
\begin{abstract}
Three dimensional electron back-scattered diffraction (EBSD) microscopy is a critical tool in many applications in materials science, yet its data quality can fluctuate greatly during the arduous collection process, particularly via serial-sectioning. Fortunately, 3D EBSD data is inherently sequential, opening up the opportunity to use transformers, state-of-the-art deep learning architectures that have made breakthroughs in a plethora of domains, for data processing and recovery. To be more robust to errors and accelerate this 3D EBSD data collection, we introduce a two step method that recovers missing slices in an 3D EBSD volume, using an efficient transformer model and a projection algorithm to process the transformer's outputs. Overcoming the computational and practical hurdles of deep learning with scarce high dimensional data, we train this model using only synthetic 3D EBSD data with self-supervision and obtain superior recovery accuracy on real 3D EBSD data, compared to existing methods.
\end{abstract}

%% file: introduction.tex
\section*{Introduction}
\label{sec:intro}
% The introduction should not include subheadings.
% \thispagestyle{dist}

Experimental methodologies for three dimensional tomography of the internal microstructure of materials has been refined considerably in the past few decades \cite{uchic_automated_2016, chapman2021afrl, polonsky_scan_2022, polonsky_three-dimensional_2019, jolley_application_2021, nguyen_alignment_2021, kotula2006tomographic, calcagnotto2010orientation}, growing to include a variety of modalities such as X-ray computed tomography (CT), optical imaging, electron imaging, energy dispersive X-ray spectroscopy (EDS), and electron back-scattered diffraction (EBSD) data, among others. In all of these cases, a volume of material is interrogated slice by slice, and these slices are then stacked and reconstructed using a variety of software tools. Three dimensional microstructure data, while often laborious to generate compared to collecting a single 2D section, can provide unique insight, including things like the topology of microstructural features like grains, pores, precipitates and the true shape and size distributions of such features. Properties such as fatigue and oxide transport are sensitive to the three dimensional arrangement of these features \cite{naragani2017investigation, sandgren2016characterization, wilson2006three}. And novel manufacturing processes like additive manufacturing also demand three dimensional characterization of microstructure to fully link the processing to the structure \cite{polonsky_scan_2022, teferra_optimizing_2021}. 

While advancements have been made in robustness and closed loop collection of data \cite{chapman2021afrl}, corruptions in data can still happen. This may not be as significant of a problem in non-destructive techniques where the data can be recollected, but in destructive techniques like serial sectioning, it may often be the case that a few slices out of a thousand have no data or are of very poor quality, lowering the accuracy of the reconstruction. The corruptions can happen for a number of reasons including removing more material (via mechanical polishing or laser/ion abalation) than planned, the electron source nearing the end of its life resulting in low signal images, the magnification on the microscope being incorrect for a slice, or the brightness/contrast settings result in an over/under saturated image. While adjustments to the control software can be made to prevent these and other issues, it is hard to a-priori imagine every reason a slice or subset of data can be corrupted. 

However, if most of the data was collected properly, we hypothesize it should be possible to infer a substantial amount of the missing data. The current common practice is to fill in missing slices is to just copy the layer above or below the missing slice \cite{chapman2021afrl}. This is reasonable in most serial sectioning cases where the slices are being collected at a high enough frequency that most of the data stays the same from slice to slice. Still, there is some room for improvement on nearest neighbor replacement type approaches, and the transformer model, which has been used to fill in text data, is a tantalizing framework for also filling in missing data in sequential image data. 

Since its introduction \cite{vaswani2017attention}, transformers have been the backbone of many impressive breakthroughs in natural language processing (NLP). In their general form, transformers are excellent at processing sequential information in parallel. Consequently, its use has spread across multiple domains, such as in computer vision \cite{khan2022transformers}, speech processing \cite{latif2023transformers}, and bioinformatics \cite{zhang2023applications}. In our case, transformers are appealing to EBSD data because the readings are inherently sequential as it represents a real physical structure.

Early large transformer models such as BERT \cite{devlin2018bert} and GPT \cite{radford2018improving} were pretrained on copious amounts of text data with the intention of learning features to model the language through associations between words and phrases with a collection of self-supervised tasks \cite{kalyan2021ammus}. Unlike supervised tasks, such as classification, which requires labeled datasets, self-supervised tasks involve automatically generating labels from unlabeled data, such as unscrambling a sequence of shuffled sentences, which force the model to vaguely learn the structure of the data. This technique remained fairly intact even when scaling to billion-parameter large language models (LLMs), enormous models that process much longer sequences over the course of hundreds of billions of tokens (language units comprising of a sequence of characters) \cite{zhang2022opt, brown2020language, touvron2023llama}. Taking inspiration from a similar popular self-supervised task in NLP, our training procedure involves randomly masking a slice in an EBSD volume and having the model predict the masked slice. 

Our slice recovery task is closely related to that of masked image modeling, masked autoencoders, and inpainting in computer vision \cite{pathak2016context, he2022masked, kong2023understanding, chang2019free, liu2021decoupled}. Even though these may be applicable and there exists plenty of transformers in computer vision \cite{dosovitskiy2020image, khan2022transformers}, we want our model to also be computationally efficient as we scale our method and leverage the fact that EBSD volumes have sparse structures \cite{dong2023deep}, the dynamics of which operate differently from typical images or videos. Regarding efficiency, the fact that EBSD produces high dimensional data means the final model's computational footprint cannot be ignored.

Taking these into consideration, our contributions are the following:
\begin{enumerate}
    \item We propose a novel method to recover missing EBSD data consisting of a scalable transformer model and straightforward projection algorithm that produces superior results compared to existing methods. This is accentuated when zoning in on the recovery accuracy of grain boundaries where other methods appear to perform poorly.
    \item We demonstrate that despite being trained solely on synthetic data, our transformer can generalize to real EBSD data \textit{without} additional training while still outperforming all the baselines. This robustness to out-of-distribution EBSD data overcomes a major limitation of relatively low amount of available real EBSD data that deep learning requires.
    \item Our results suggest that serial sectioning experiments using similar experimental parameters to those in our test datasets, collection time can effectively be slashed by up to 25\% with very little error in predicted voxel orientation, as the collection of every fourth slice can be bypassed. With the current lengthy procedure of EBSD, this is a significant improvement in efficiency.
\end{enumerate}
% Throughout this paper, there are data structures that have identical shapes with the exception of dimensions that do no exist which are grouped toward the end. Such is the case for $\bcA \in \RR^{N_1 \times \cdots \times N_n}$ and $\bcB \in \RR^{N_1 \times \cdots \times N_k}$ for some $1\leq k < n$. For brevity, we refer the shape of both volumes simultaneously with $(N_1, \dots, N_k, *)$, where $*$ captures all remaining dimensions.

%% file: background.tex
\section*{Background}
\label{sec:background}

\subsection*{Terminology and Notation}

For this paper, we use bold capital letters (e.g. $\bA$) for matrices and capital calligraphic letters (e.g. $\bcA$) for tensors. Scalars are represented by plain uppercase and lowercase letters (e.g. $A$ and $a$). Tuples are used to describe the shape of a structure, so if $\bcA \in \RR^{N_1 \times N_2 \times N_3}$, then $\bcA$ is of shape $(N_1, N_2, N_3)$ and vice versa. A dimension is an index (starting at 1) in the shape tuple, and the size of a dimension $i$ is the value of the shape tuple at that index $i$. When we refer to multiple volumes that share the same first $k$ dimensions (e.g. $\bcA \in \RR^{N_1 \times N_2}$, $\bcB \in \RR^{N_1 \times N_2 \times N_3}$, and $\bcC \in \RR^{N_1 \times N_2 \times N_3'}$), we use $*$ to capture all remaining and possibly different dimensions (e.g. $\bcA$, $\bcB$, and $\bcC$ have shape $(N_1, N_2, *)$).

% (e.g. $\bcA \in \RR^{N_1 \times \cdots \times N_m}$ and $\bcB \in \RR^{N_1 \times \cdots \times N_k \times N_{k+1}' \times \cdots \times N_n'}$), we use $*$ to capture all remaining and possibly different dimensions (e.g. $\bcA$ and $\bcB$ have shape $(N_1, \dots, N_k, *)$).

% \subsection*{Electron Backscatter Diffraction Microscopy}

\subsection*{Transformers}

In this section, we outline the technical details of the encoder-only transformer based on its first introduction \cite{vaswani2017attention}. The same source describes encoder-decoder and decoder-only transformers for interested readers.

\subsubsection*{Input Processing}

The first layer of a transformer is the embedding layer. For a single length $N$ sequence of $D_{in}$-dimensional vectors, $\bX\in \RR^{N \times D_{in}}$, each vector is transformed into a vector of size $D_{out}$. If the input is a sequence of tokens such as in language tasks, then $D_{in} = 1$. The transformation could be a linear one, or a more complex nonlinear one such as another neural network. The result is an embedded sequence of shape $(N, D_{out})$. Next, positional encoding is added to the embedding to inject positional information to each vector in the sequence. Positional encodings can either be fixed or learned.

% The input to a transformer can take different forms which can affect how the embedding step operates. For a single length $N$ sequence of tokens, $\bX\in \RR^{N \times 1}$, each token is embedded into a vector of length $D_{out}$. For a single length $N$ sequence of $D_{in}$-dimensional vectors, $\bX\in \RR^{N \times D_{in}}$, each vector is transformed into a vector of length $D_{out}$. This could be a linear transformation or something more complex such as another neural network. Regardless of the input type, the result is an embedded sequence of shape $(N, D_{out})$.

\subsubsection*{Multi-head Self-attention}

The idea of self-attention is to find varying degrees of associations between elements in a sequence. Transformers typically implement multi-headed self-attention within each encoder layer, which allows the model to learn different types of associations. Let $\bX \in \RR^{N \times D}$ be the input sequence into the attention mechanism and $\bW_h^Q, \bW_h^K, \bW_h^V \in \RR^{D \times \frac{D}{H}}$ for head $h = 1, 2, \dots, H$. Additionally, let $\sigma$ be the row-wise softmax operator. Then, the attention for head $h$, $\bA_h$ is defined as
\begin{align}\label{eq:attn}
    \bA_h = \sigma\left(\frac{\bQ_h \bK_h^\top}{\sqrt{D / H}} \right) \bV_h,
\end{align}
where $\bQ_h = \bX \bW_h^Q, \bK_h = \bX \bW_h^K, \bV_h = \bX \bW_h^V$ are the query, key, and value matrices, respectively. Next, the results are concatenated and linearly transformed to produce the output $\bY = [\bA_1 \cdots \bA_H] \bW_O$ for $\bW_O \in \RR^{D \times D}$.
Note that $\bA_1, \dots, \bA_H$ can be computed in parallel. 

The main source of inefficiency is the computation of $\bQ_h \bK_h^\top$ which levies a computational and memory cost of $\cO (N^2)$. Consequently, the utility of transformers can be limited for long sequences without proper hardware. This becomes a dire issue as the dimensionality of the input grows, such as in images and videos which can be flattened into a long sequence. Thankfully, there has been substantial work in reducing the complexity with more efficient attention approximation mechanisms, such as Linformer \cite{wang2020linformer}, Reformer \cite{kitaev2020reformer}, 
% Nystr\"omformer \cite{xiong2021nystromformer}, 
Big Bird \cite{zaheer2020big}, and many others \cite{tay2022efficient}.

\subsubsection*{Axial Self-attention}

% Our missing slice recovery task can be framed as a computer vision task, and there exists a plethora of transformers for vision \cite{khan2022transformers}. While it is possible to use them directly, scaling the input size greatly increases the amount of computation required due to the large dimensionality of our data. 
When it comes to high-dimensional data, such as the EBSD volumes, the sequence length explodes if we flatten (or patchify in the case of many transformers in computer vision \cite{khan2022transformers}) the input and apply vanilla self-attention, making attention computation a serious bottleneck. Furthermore, since we observe a lot of structure with EBSD data, it is reasonable to utilize some simplified attention mechanism. Our model uses axial attention \cite{ho2019axial, wang2020axial} which runs the self-attention mechanism along the dimensions of the input tensor. For instance, in a cube, each voxel only attends to voxels in the same row, column, or depth. This greatly reduces the amount of computation and memory, especially for higher order tensors. Intuitively, axial attention is appropriate because we hypothesize that since local information can be quite uniform (nearby voxels are likely to be in the same grain), long range information should also be included. With axial attention, it is highly likely to also obtain information in other grains and its boundaries.

In terms of implementation, the formula for multi-headed attention \eqref{eq:attn} can be reused. Let $\bcX \in \RR^{N_1, \dots, N_K, D}$ be a single $K$-dimensional input where $(N_1, \dots, N_K)$ defines the shape of the volume and $D$ is the embedding size. As an example, suppose we are interested in finding axial attention along the $k$-th dimension for $1<k<K$. With proper dimension permutation and flattening, we can reshape $\bcX$ to a tensor of shape $(\prod_{i\neq k} N_i, N_k, D)$ and compute multi-headed attention by treating the first dimension as the batches. In our model, we repeatedly use the outputs of axial attention to compute axial attention along the next dimension.

\subsubsection*{Feedforward Blocks}

The next major component to create a transformer is the feedforward block, a multilayer perceptron with a single hidden layer. The activation function is often either a Gaussian Error Linear Unit (GELU) or Rectified Linear Unit (ReLU) function, and although the input and output sizes are identical, the hidden size can be much larger. Consequently, although self-attention is a computational bottleneck, the feedforward blocks are the dominant source of parameters in the model.

\subsubsection*{Putting it All Together}

\begin{figure}
    \centering
    \includegraphics[width=0.55\linewidth]{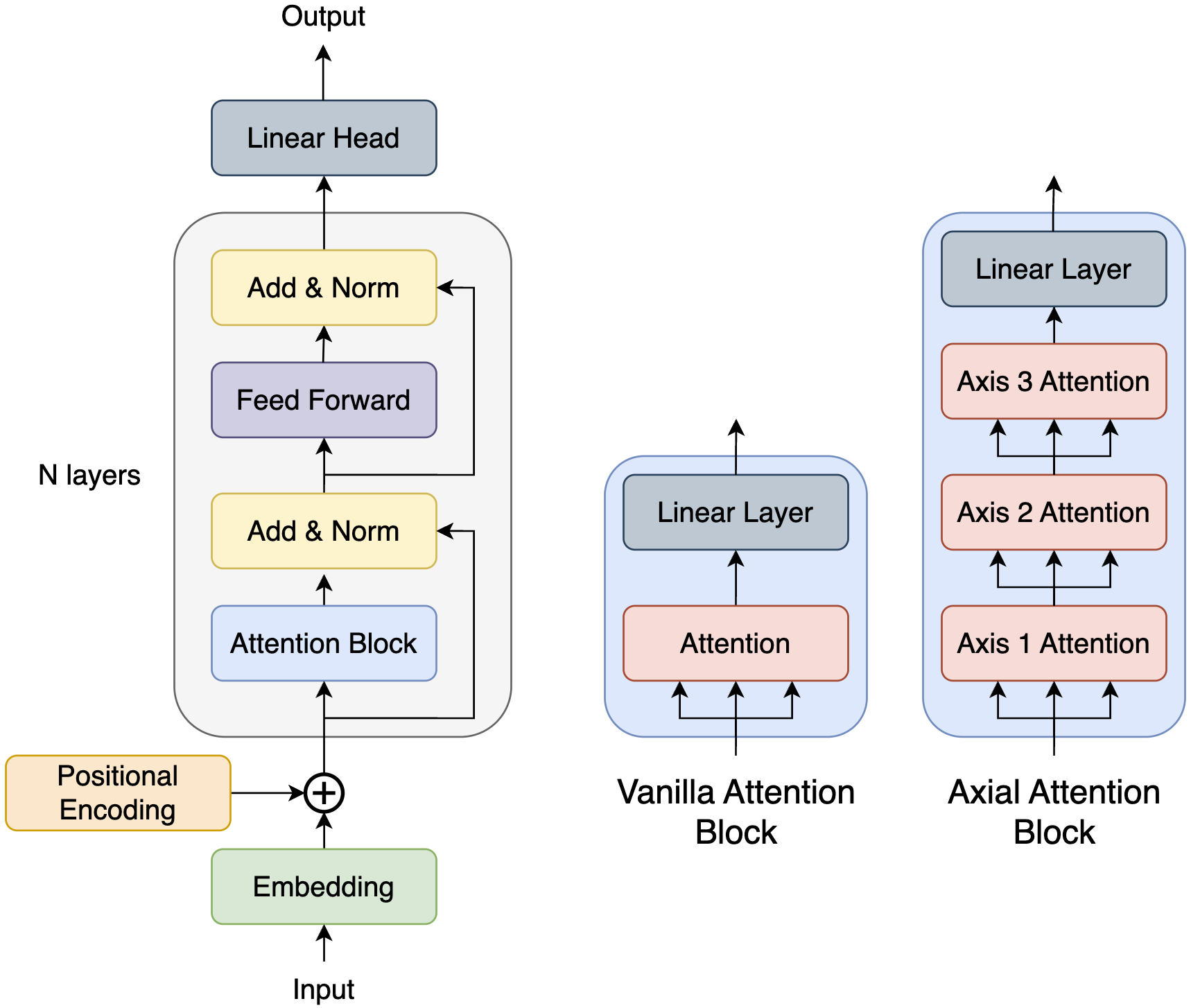}
    \caption{Transformer architecture (left) with either vanilla attention (center) or axial attention (right). The axial transformer is produced by simply substituting full attention with axial attention. Although axial attention has significantly more layers, it scales much more favorably for high-dimensional data. Attention layers take three inputs for the query, key, and value.}
    % \caption{Architecture of our axial transformer and axial attention block. This would be a vanilla encoder-only transformer if the axial attention blocks are replaced with normal self-attention.}
    \label{fig:arch}
\end{figure}

With all these components, we are ready to define the general architecture of a transformer. First, an input is processed by an embedding layer and positional encoding layers at the start. Then, the input progresses into the encoder, a sequence of alternating self-attention and feedforward blocks. For training stability, residual connections and normalization layers are inserted in between each attention block and feedforward block. Finally, the features from the encoder are linearly projected to be the desired shape of the output. A visualization of a vanilla and axial transformer, which mainly differ in their attention implementation, is depicted in Figure \ref{fig:arch}.

%% file: method.tex
\section*{Method}
\label{sec:model}

We propose a transformer model to learn missing slices of EBSD data, followed by a projection step to smooth out the voxel values. Due to limited real EBSD data, our goal is to train this model on a large and diverse synthetic dataset and demonstrate that it can generalize to real EBSD data, which we evaluate on two nickel superalloy EBSD volumes, one for alloy IN625 \cite{menasche2021afrl, shade2019afrl} and one for alloy IN718 \cite{stinville2022multi, stinville2022multidataset}. The following repository contains the code for our method:
\url{https://github.com/hdong920/ebsd_slice_recovery}.

\subsection*{Data Description and Preparation}\label{data_description}

\begin{figure}[h]
    \centering
    \begin{subfigure}[b]{0.4\linewidth}
         \centering
         \includegraphics[width=1\linewidth]{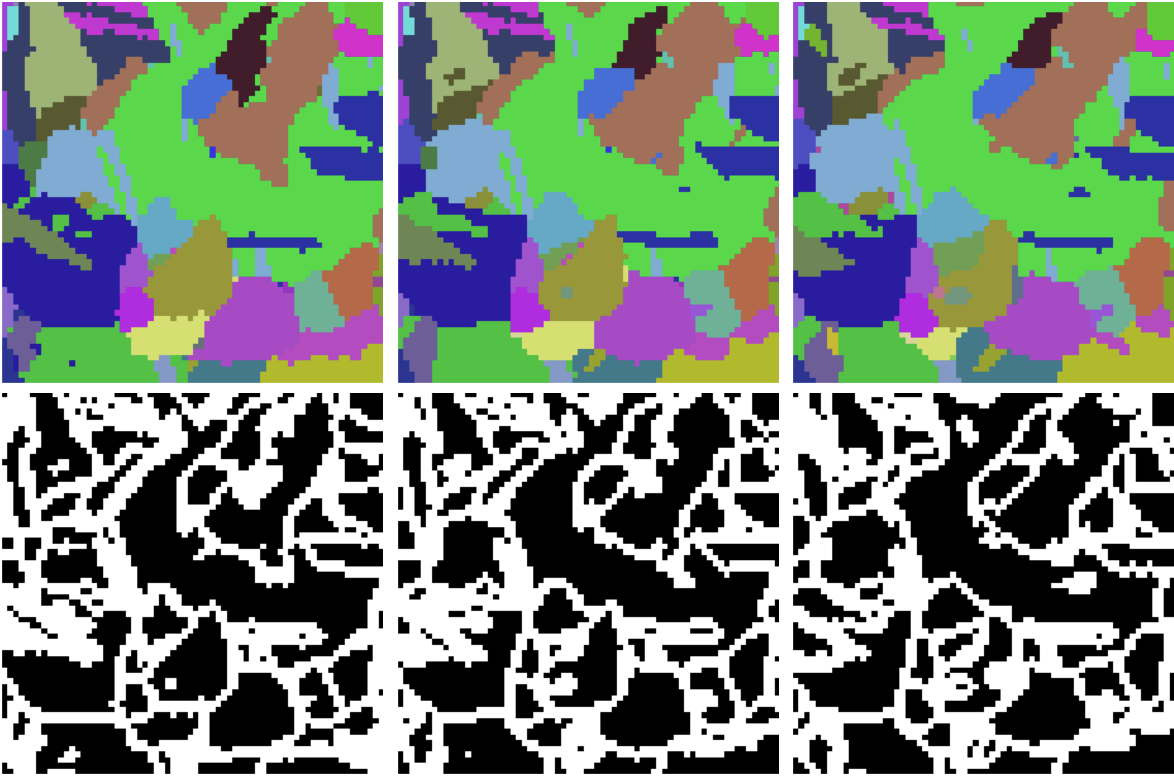}\\
         \caption{}
        \label{fig:in625_examples}
     \end{subfigure}
     \hspace{0.06\linewidth}
     \begin{subfigure}[b]{0.4\linewidth}
         \centering
         \includegraphics[width=1\linewidth]{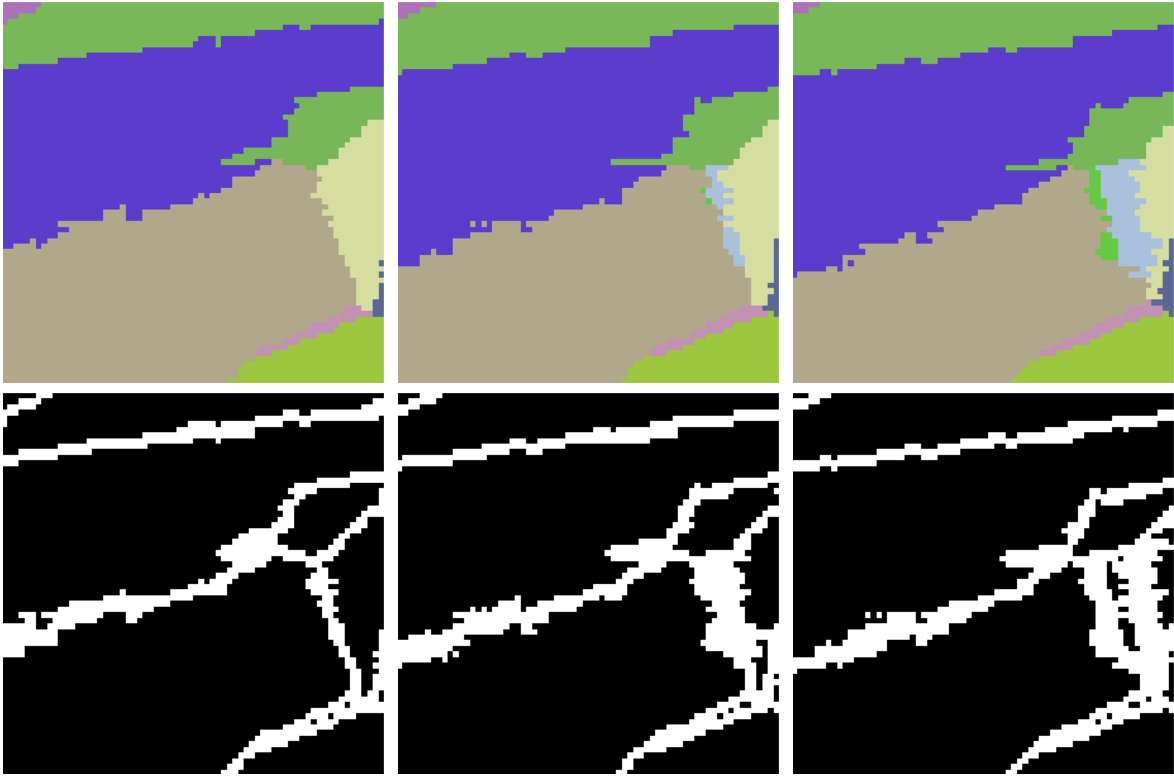}\\
         \caption{}
        \label{fig:in718_examples}
     \end{subfigure}
    \caption{Three consecutive crops of real EBSD slices from IN625 \cite{chapman2021afrl} (Figure \ref{fig:in625_examples}) and IN718 \cite{stinville2022multi} (Figure \ref{fig:in718_examples}). The top row shows the cubochoric values that are scaled and shifted for better visualization, and the bottom row shows the same region in \texttt{Boundaries}. Note that the height and width of each slice here is 64 voxels and not the original shape.}
    \label{fig:ebsd_examples}
\end{figure}

Each dataset (both synthetic and experimental) includes orientation information at every voxel in a 3D image. For the experimental data, a substatial amount of preprocessing was done to handle the alignment of the data and clean up noise; a complete description of the preprocessing may be found in Chapman et al.\cite{chapman2021afrl} and Stinville et al.\cite{stinville2022multi} In particular, we remove any grains smaller than 27 voxels ($3^3$) and average orienations per grain. The original volume containing Euler angles is of shape $(N_1, N_2, N_3, 3)$, where $N_1, N_2, N_3$ are the physical dimensions, and the final dimension represents the 3 Euler angles needed to define an orientation. Additional arrays are then computed from the input data:

\begin{itemize}
    \item \texttt{Cubochorics}: A volume of shape $(N_1, N_2, N_3, 3)$ where the last dimension contains the cubochoric coordinates \cite{rocsca2014new} converted from the original Euler angles at each voxel. Cubochoric coordinates are chosesn since the Euclidean metric is used for regressing the transfomer model. The Euclidean distance between points in Euler angle space does not necessarily relate to the angular distance between those points. While this is also true for points in cubochoric space, as the cubochoric representation is an equal-volume mapping of \emph{SO}(3) onto a grid as opposed to an equal-angle mapping, in practice the angular discretization between Euclidean neighbors in cubochoric space is relatively consistent across the grid. Thus, for all of our experiments, we operate completely in cubochoric space, under the assumption that the Euclidean metric is a reasonable approximation for similarity between points in this space.
    \item \texttt{IDs}: A volume of shape $(N_1, N_2, N_3)$ which assigns identification numbers (IDs) to denote which grain each voxel belongs to. Each ID number has a unique vector of cubochoric coordinates associated with it. While not needed during training, these ID numbers will be used to smooth our model outputs and evaluate model accuracy. For experimental data, the IDs are found by segmenting the grains using a misorientation tolerance \cite{chapman2021afrl, stinville2022multi}, and for synthetic data, the IDs are generated alongside the orientation data. 
    \item \texttt{Boundaries}: A volume of shape $(N_1, N_2, N_3)$ containing Boolean values indicating if a voxel is on the grain boundary. More specifically, a voxel is a boundary voxel is at least one of its neighbors is of a different ID number than itself, where two voxels are neighbors if they share a face.
\end{itemize}
To illustrate, Figure \ref{fig:ebsd_examples} contains example slices of (scaled and shifted) \texttt{Cubochorics} and \texttt{Boundaries} of IN625 and IN718.

\begin{figure}
    \centering
    \begin{subfigure}[b]{0.15\linewidth}
         \centering
         \includegraphics[width=1\linewidth]{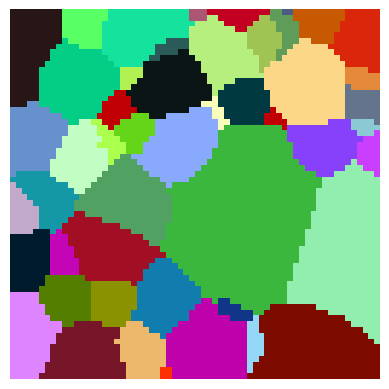}\\
         \caption{}
        \label{fig:synthetic_exmaple_twins0}
     \end{subfigure}
     \begin{subfigure}[b]{0.15\linewidth}
         \centering
         \includegraphics[width=1\linewidth]{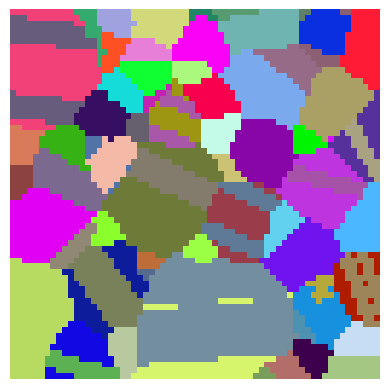}\\
         \caption{}
        \label{fig:synthetic_exmaple_twins2}
     \end{subfigure}
     \begin{subfigure}[b]{0.15\linewidth}
         \centering
         \includegraphics[width=1\linewidth]{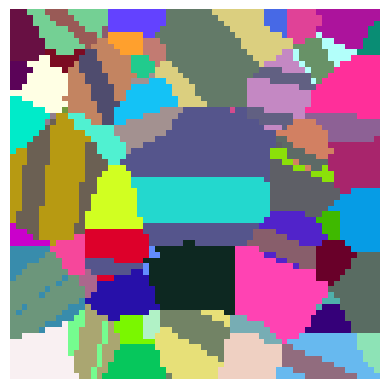}\\
         \caption{}
        \label{fig:synthetic_exmaple_twins4}
     \end{subfigure}
     \begin{subfigure}[b]{0.15\linewidth}
         \centering
         \includegraphics[width=1\linewidth]{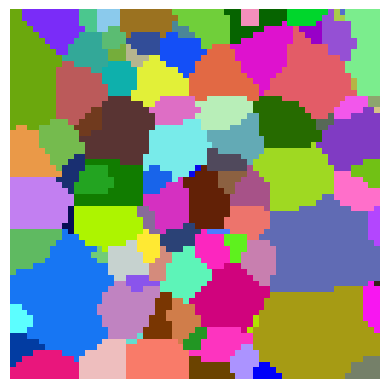}\\
         \caption{}
        \label{fig:synthetic_exmaple_size2}
     \end{subfigure}
     \begin{subfigure}[b]{0.15\linewidth}
         \centering
         \includegraphics[width=1\linewidth]{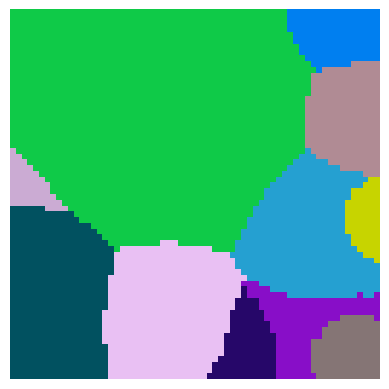}\\
         \caption{}
        \label{fig:synthetic_exmaple_size3}
     \end{subfigure}
    \caption{Example synthetic slices of cubochoric values that are scaled and shifted for visualization. Keeping mean grain size fixed, Figures \ref{fig:synthetic_exmaple_twins0}, \ref{fig:synthetic_exmaple_twins2}, and \ref{fig:synthetic_exmaple_twins4} show generated slices when we specify the mean frequency of twins per grain to be 0, 2, and 4, respectively. With no twins, Figures \ref{fig:synthetic_exmaple_size2} and \ref{fig:synthetic_exmaple_size3} show generated slices when we specify the mean grain size to be 2 and 3, respectively. All slices have a height and width of 64 voxels.}
        \label{fig:synthetic_examples}
\end{figure}

Synthetic volumes are generated via DREAM.3D \cite{groeber2014dream}. Each synthetic training and validation volume is of shape $(192, 192, 192, *)$ and $(64, 192, 64, *)$, respectively. Within DREAM.3D, we generate 9 training volumes while independently varying the mean grain size and mean transformations per grain. In particular, we generate volumes with mean grain sizes 2, 2.5, and 3 with no twins; we also generate volumes with mean twins frequencies 0, 1, 2, 3, 4, and 5 while fixing mean grain size to be 2.3. Due to the nature of the software, these parameters are unitless, as we can always synthetically increase or decrease the granularity of the volume. For example slices of these synthetic volumes, see Figure \ref{fig:synthetic_examples}. The grain size distributions for each dataset are shown in Figure \ref{fig:synthetic_examples} as probability plots. The natural logarithm of the normalized grain sizes are shown; ideal lognormally distributed data would lie on straight lines in such plots. In general, the grain sizes are primarily lognormal near their means, with noted deviations from lognormality in their tails, which is a known phenomenon \cite{donegan2013ev}.

\begin{figure}
    \centering
    \begin{subfigure}[b]{0.4\linewidth}
         \centering
         \includegraphics[width=1\linewidth]{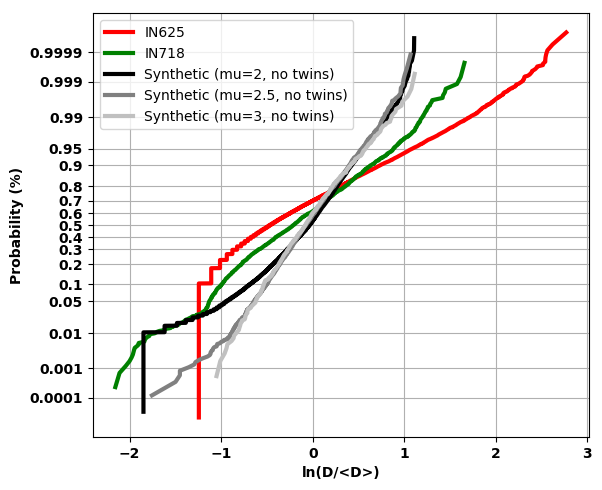}\\
         \caption{}
        \label{fig:prob_plot_notwins}
     \end{subfigure}
     \begin{subfigure}[b]{0.4\linewidth}
         \centering
         \includegraphics[width=1\linewidth]{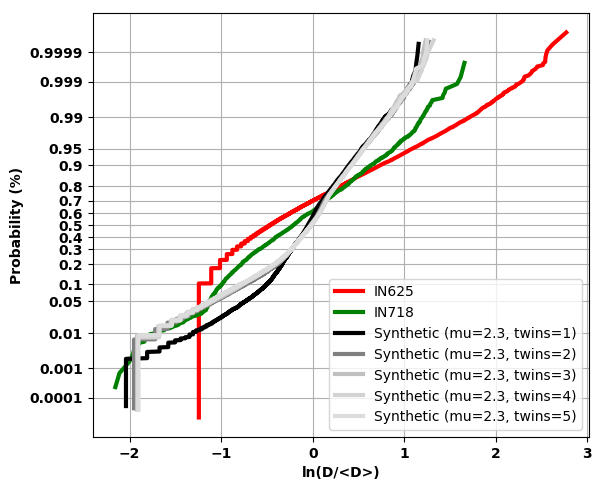}\\
         \caption{}
        \label{fig:prob_plot_twins}
     \end{subfigure}
         \caption{Probablity plots showing the grain size distributions for each dataset. Figure \ref{fig:prob_plot_notwins} compares the real test datasets to the synthetic training datasets without twins, while Figure \ref{fig:prob_plot_twins} compares the real test datasets to the synthetic training datasets with twins. Grain sizes are represented as sphere equivalent diameters, D, normalized by the distribution mean.}
        \label{fig:prob_plots}
\end{figure}

\subsection*{Training Details}

We first describe a few data augmentation steps. The number of unique cubochoric coordinates is quite limited, so color shift transformations, along with other augmentations, are critical. \textit{Note that use of the word "color" is loose here because we treat the cubochoric coordinates the same as color channels in computer vision.} In particular, our augmentations include random linear color shifts, rotations, and flips.

Using the 9 synthetic volumes generated by DREAM.3D, we train our axial transformer model in a self-supervised fashion similar to masked language modeling tasks \cite{devlin2018bert}. These volumes are first normalized along each of the three cubochoric indices. Each training input is sampled from a randomly chosen volume with physical dimensions randomly permuted. Not only is cropping necessary due to computational limits, it also acts as another form of augmentation. For a sample $\bcX_\star \in \RR^{64\times 7 \times 64 \times 3}$, one of the central 5 slices (along the second dimension) is randomly masked. If $m$ is the index of the masked or unobserved slice, define $\bcM \in {[0,1]}^{64\times 7 \times 64 \times 3}$ to be a mask such that $[\bcM]_{\cdot, m} = 0$ and $1$ elsewhere. Then, the model input and output will be $\bcX_\star \odot \bcM$ and $\widehat{\bcX} \in \RR^{64\times 7 \times 64 \times 3}$, respectively.

\begin{figure}
    \centering
    \includegraphics[width=0.3\linewidth]{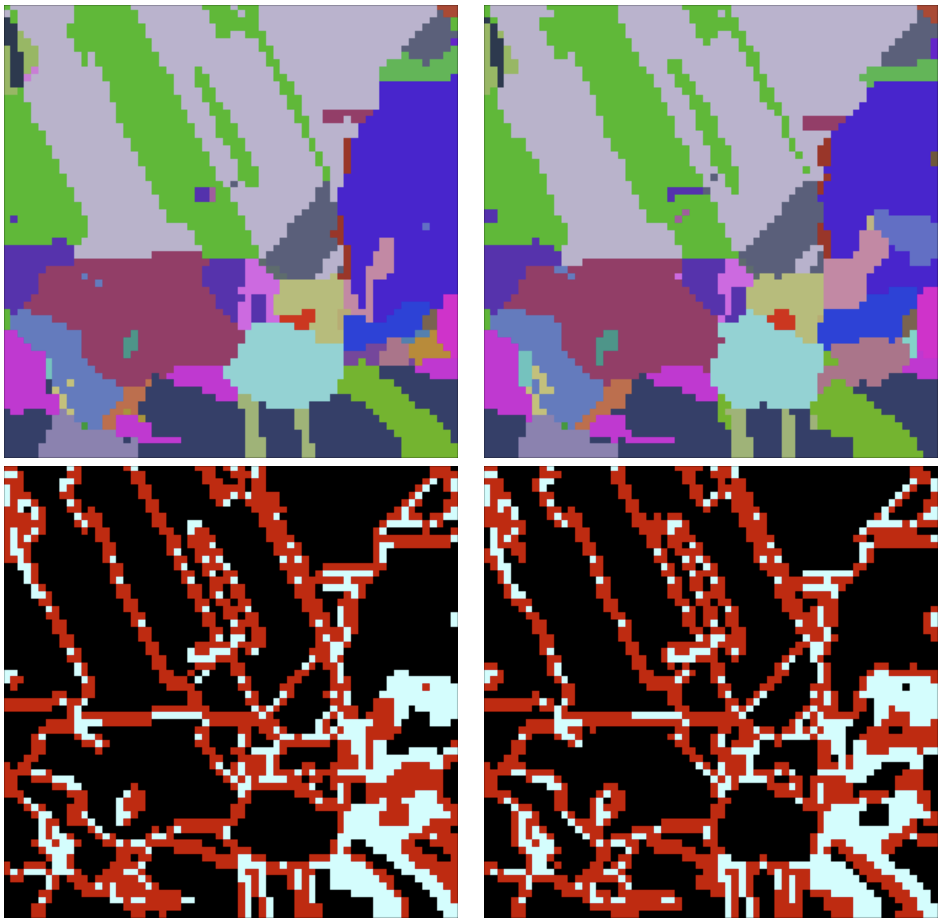}
    \caption{Changes between slices are captured in boundary voxels. The top row shows two consecutive slices. The bottom row shows their respective slices in \texttt{Boundaries} (red) with the ID difference between the two slices overlaid (white). Note that differences between these slices lie fully in the boundary voxels of both slices.}
    \label{fig:differences}
\end{figure}

Since EBSD produces very structured data, we can leverage this to design a more effective loss function. A simple mean-squared-error (MSE) loss function across all voxels would not be sufficient because most voxels in the input are observed, so we would risk learning an identity function which would produce a fairly low loss value. A better approach would be to find the MSE of only the missing slice, analogous to masked language modeling objective functions \cite{devlin2018bert}. However, based on Figure \ref{fig:ebsd_examples}, we see that many of voxels remain unchanged across multiple slices, and the slice-to-slice changes all lie along the grain boundaries (see Figure \ref{fig:differences}). Since voxels within grains are easy to predict the values of, the source of difficulty is recovering the boundary voxels. Therefore, we opt for a MSE loss function that only considers boundary voxels in the missing slice. More formally, letting $k$ be the index of the missing slice and $\bcE \in {[0,1]}^{64 \times 64 \times 1}$ be the missing slice's boundaries map in \texttt{Boundaries}, the loss function we use is 
\begin{align} \label{loss}
    \cL(\widehat{\bcX}, \bcX_\star) = \frac{\left\| [\widehat{\bcX} - \bcX_\star]_{\cdot, m} \odot \bcE \right\|_{\fro}^2}{\left\|\bcE \right\|_0},
\end{align}
following broadcasting rules. By the way we defined $\bcE$, $\left\|\bcE\right\|_0$ is the number of unobserved boundary voxels. In summary, this loss function essentially averages MSE error across all unobserved boundary voxels.

% More formally, let $\bcE \in {[0,1]}^{64 \times 7 \times 64 \times 1}$ be the corresponding crop of \texttt{Boundaries} based on the location of $\bcX_\star$. Then, with the model output $\widehat{\bcX}$, the loss function we use is 
% \begin{align}
%     \cL(\bcX_\star, \widehat{\bcX}) = \frac{\left\| (\bcX_\star - \widehat{\bcX}) \odot \bcM \odot \bcE \right\|_{\fro}^2}{\left\|\bcM \odot \bcE \right\|_1},
% \end{align}
% again following broadcasting rules. By the way we defined $\bcM$ and $\bcE$,  $\left\|\bcM \odot \bcE\right\|_1 = \left\|\bcM \odot \bcE\right\|_0$ and is just the number of unobserved boundary voxels. In summary, this loss function essentially averages error across all unobserved boundary voxels.

Using this scheme, we train our 8-headed 8-layer model using stochastic gradient descent with a momentum parameter of 0.9 and weight decay of $1\mathrm{e}{-5}$. Using cosine schedules, we warm-up the learning rate up to $0.01$ for 8000 steps. While we decay the learning rate until 160000 total gradient steps are taken with batches of 1 sample, performance plateaus around halfway. The model uses learnable positional encoding, 4 attention heads, an embedding size of 128, and a feedforward size of 512. The feedforward block is consists of two 3D convolutions with a window size of 3 along each dimension separated by a GELU. See Figure \ref{fig:arch} for a visualization of the architecture. Notably, our model is compact for a transformer, only consisting of slightly under 30 million parameters. We also apply 10\% dropout. Recall that the transformer returns predicted cubochoric coordinates at each voxel of the same shape as the input, but only the masked slice contributes to the loss function \eqref{loss}.

% \subsection*{Prediction}
\subsection*{Nearest Neighbor Projection}

\begin{figure}
    \centering
    \includegraphics[width=1\linewidth]{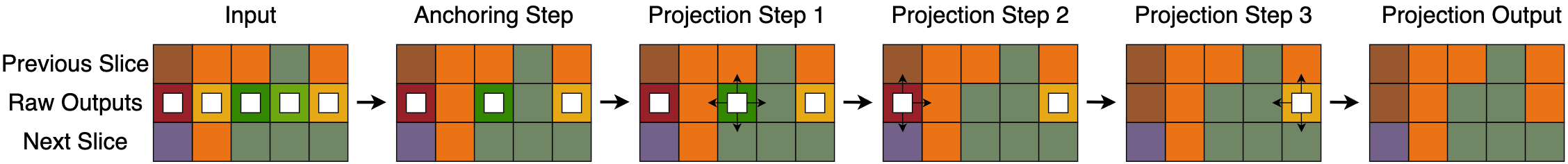}
    \caption{Projection algorithm to take transformer outputs and assign voxels to grains. White boxes denote which voxels have not been projected yet. The input contains the transformer output and observed adjacent slices. First, the anchoring step assigns voxels whose neighbors in adjacent slices are from the same grain. Then, voxels are sequentially projected to their neighbors, as indicated by the small arrows, starting with the voxels with the most observed and previously projected neighboring voxels.}
    \label{fig:projection}
\end{figure}

The final step is to use outputs of our transformer, which are continuous values, to assign each voxel to a grain and produce a smoother slice with fewer intra-grain variations. To do this, we first construct a dictionary relating observed grain IDs to cubochoric coordinates. Then, we assign voxels whose neighbors (voxels that share a face) that reside in the previous and next slice have the same ID to be that ID. We observe these voxels act like anchors that provide more neighbors for voxels that are more difficult to classify, which empirically improves the recovery. Next, we begin projecting voxels with the most observed or previously projected neighbors and incrementally decrease the requirement on the number of neighbors when further projections are impossible until all voxels have been assigned an ID. Projections are determined by the minimum $\ell_2$ distance to relevant cubochoric coordinates pulled from the dictionary based on neighboring IDs. Summarized in Figure \ref{fig:projection}, the projection algorithm essentially turns the transformer outputs of the missing slice into IDs which can then be converted into cubochoric coordinates.

% To do this and encourage continuity of grains, we project each missing voxel's predicted cubochoric coordinates to the closest cubochoric coordinates among its observed or previously projected neighbors where two voxels are neighbors if they share a face. Distance is measured by $\ell_2$ distance. With some error tolerance, we first project voxels that have identical neighbors that reside in previous and next slice to their neighbors' values. We observe this acts like an anchor that empirically improves the recovery. Next, we begin projecting voxels with the most observed or previously projected neighbors and incrementally decrease the requirement on the number of neighbors when further projections are impossible until we have projected all voxels. Note that this also maintains a one-to-one correspondence between IDs and cubochoric coordinates and between IDs and Euler angles.

%% file: experiments.tex
\section*{Experiments}
\label{sec:experiments}

With our method defined, we now evaluate its performance on synthetic and real EBSD data with no additional training or fine tuning. Even though our method overwhelmingly outperforms the baselines in terms of recovery, we also point out some weaknesses and avenues for improvement.

There are three baselines we compare to. One is k-nearest neighbors (KNN) where a voxel's ID is determined by vote based only on observed or previously assigned IDs among its neighboring voxels. This is the exact process of the projection step (including the anchoring procedure), except instead of using a distance metric, voxels are assigned IDs by a vote system. Ties are broken randomly. Another method, which is currently employed as a simple solution to missing slices, is to copy an adjacent slice to replace the missing slice. This usually maintains fairly decent accuracy since the changes from slice to slice are minuscule compared to the number of voxels. As such, copying the previous and next slice are our other two baselines. 

\subsection*{Performance}

Because the changes from slice to slice are captured by boundary voxels, we define two different accuracy metrics using the IDs. We denote the accuracy of each voxel in the recovered slice as the \textit{overall accuracy} and the accuracy only considering boundary voxels in the recovered slice as the \textit{boundary accuracy}. The latter poses a greater challenge for all models, as it is consistently lower than the overall accuracy where the influence of non-boundary voxels often dominates.

While we report the mean accuracies and standard deviation across validation samples, the performance of one method is heavily correlated with the performance of others. For instance, if a particular slice is difficult to recover for one method, it is likely to be difficult for all other methods. To better compare the performances between our transformer and each baseline, we obtain differences in accuracy for each sample. Namely, we find $d(m, \bcX_\star, \widehat{\bcX}, \widehat{\bcX}_{b}) := m(\bcX_\star, \widehat{\bcX}) - m(\bcX_\star, \widehat{\bcX}_{b})$ for an accuracy metric $m$, ground truth $\bcX_\star$, and outputs, $\widehat{\bcX}$ and $\widehat{\bcX}_b$, from our transformer and baseline $b$, respectively. Computing this across all samples, the result is a distribution of \textit{accuracy improvements}.

% We find that the performance of one method is heavily correlated with the performance of others. For instance, if a particular slice is difficult to recover for one method, it is likely to be difficult for all other methods. To better compare the performances between our transformer and each baseline, we obtain differences in performance for each sample. Namely, we find the difference metric $d(m, \bcX_\star, \widehat{\bcX}, \widehat{\bcX}_{b}) := m(\bcX_\star, \widehat{\bcX}) - m(\bcX_\star, \widehat{\bcX}_{b})$ for an error metric $m$, ground truth $\bcX_\star$, and outputs, $\widehat{\bcX}$ and $\widehat{\bcX}_b$, from our transformer and baseline $b$, respectively. Computing this across all samples, the result is a distribution of performance differences. We present these metrics alongside the average accuracy across all samples.

\subsubsection*{Synthetic Volumes}

\begin{figure}
    \begin{minipage}[b]{1\linewidth}
     \centering
     \begin{subfigure}[b]{0.49\linewidth}
         \centering
         \includegraphics[width=1\linewidth]{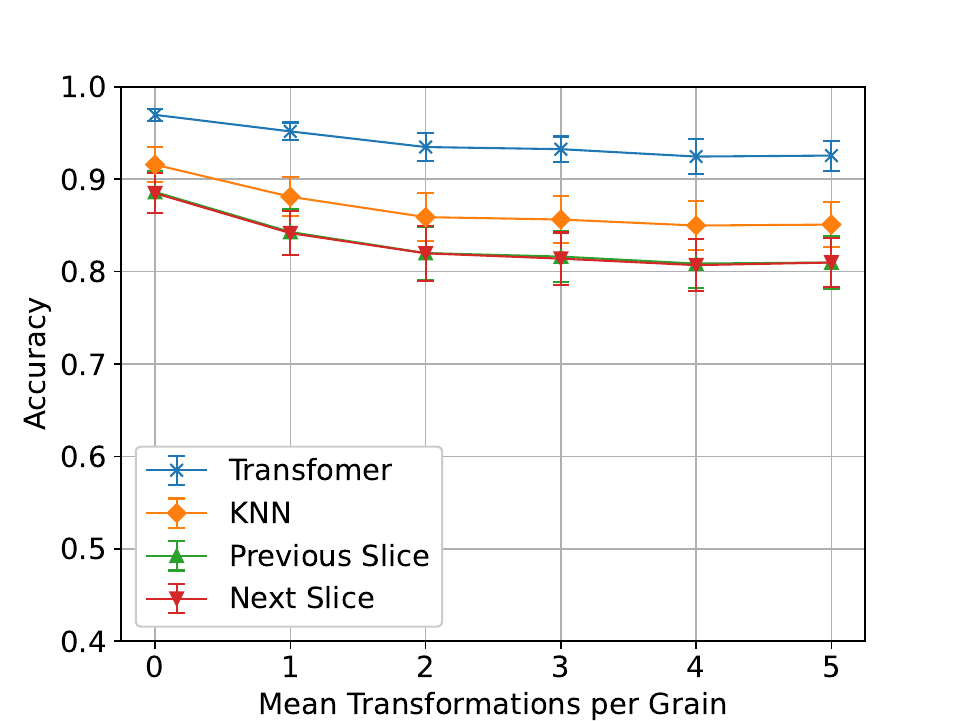}\\
         \caption{Overall Accuracy}
        \label{fig:twins_overall_acc_avg}
     \end{subfigure}
     \begin{subfigure}[b]{0.49\linewidth}
         \centering
         \includegraphics[width=1\linewidth]{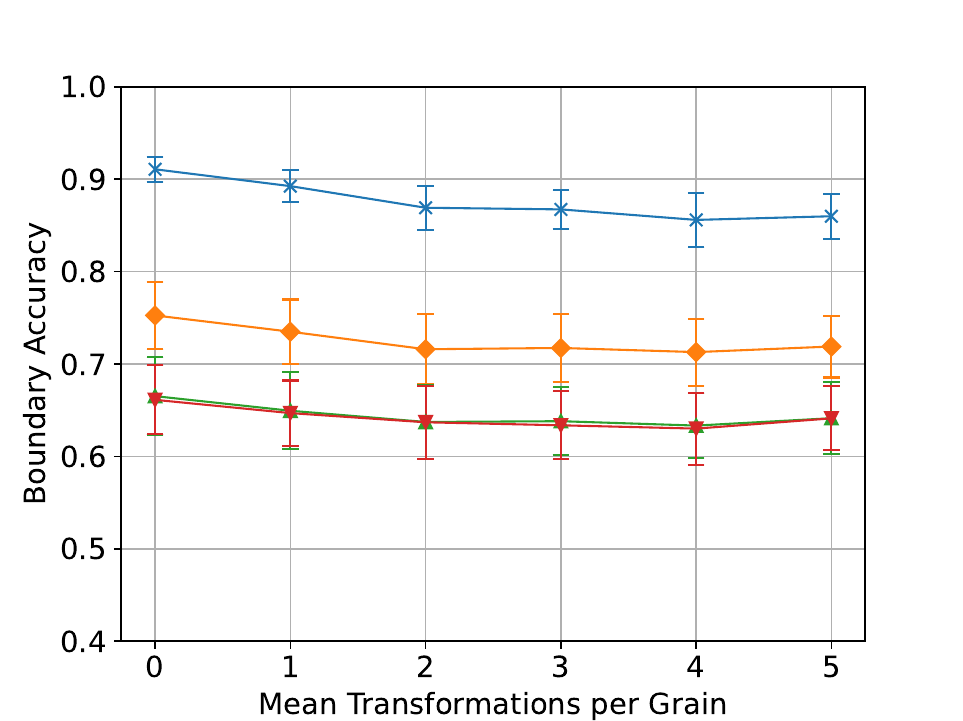} \\
         \caption{Boundary Accuracy}
        \label{fig:twins_edge_acc_avg}
     \end{subfigure}
     % \vspace{-0.1in}
     \begin{subfigure}[b]{0.49\linewidth}
         \centering
         \includegraphics[width=1\linewidth]{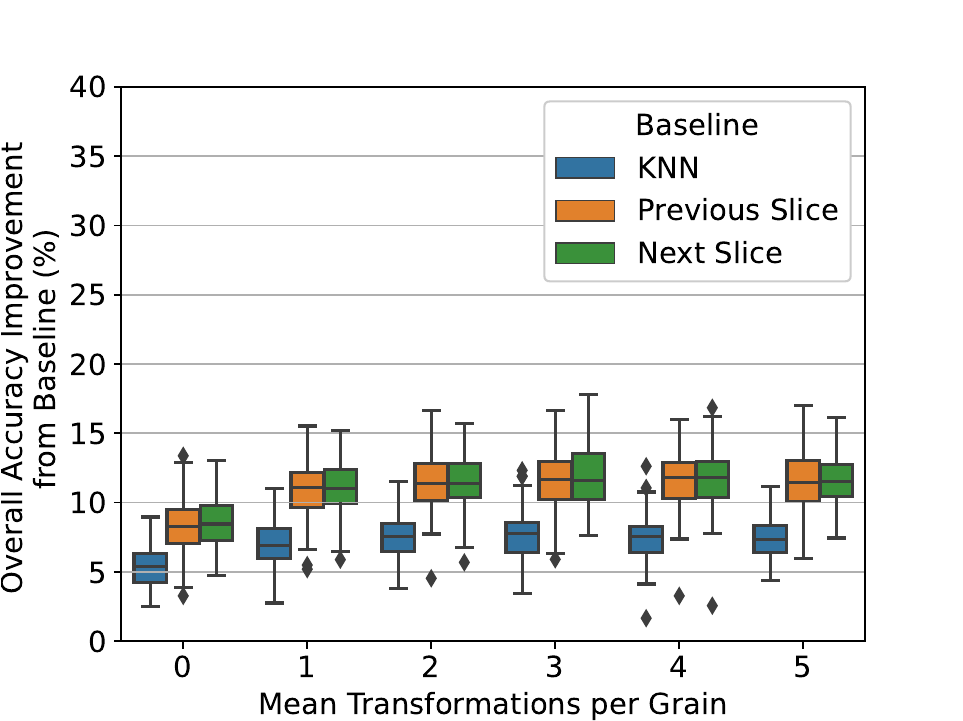}\\
         \caption{Overall Accuracy Improvements}
        \label{fig:twins_overall_acc}
     \end{subfigure}
     \begin{subfigure}[b]{0.49\linewidth}
         \centering
         \includegraphics[width=1\linewidth]{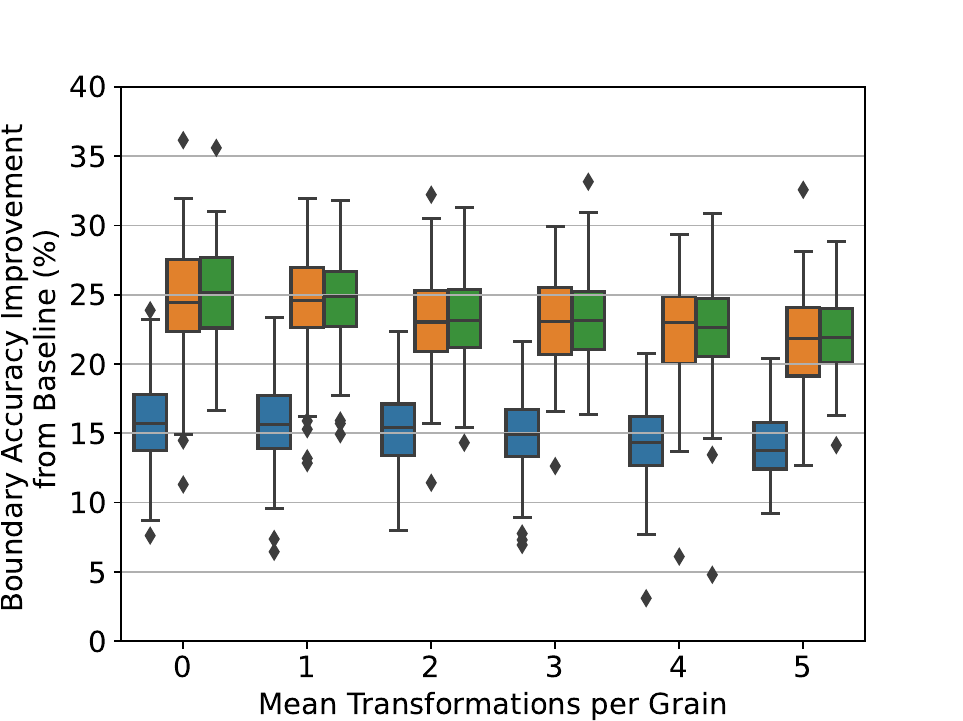} \\
         \caption{Boundary Accuracy Improvements}
        \label{fig:twins_edge_acc}
     \end{subfigure}
     \vspace{-0.1in}
     \end{minipage}
        \caption{Overall accuracy (Figure \ref{fig:twins_overall_acc_avg}), boundary accuracy (Figure \ref{fig:twins_edge_acc_avg}), overall accuracy improvements (Figure \ref{fig:twins_overall_acc}), and boundary accuracy improvements (Figure \ref{fig:twins_edge_acc}) of synthetic volumes with varying twin frequencies. Error bars in Figures \ref{fig:twins_overall_acc_avg} and \ref{fig:twins_edge_acc_avg} represent one standard deviation.}
        \label{fig:twins_acc}
\end{figure}

\begin{figure}
    \begin{minipage}[b]{1\linewidth}
     \centering
     \begin{subfigure}[b]{0.49\linewidth}
         \centering
         \includegraphics[width=1\linewidth]{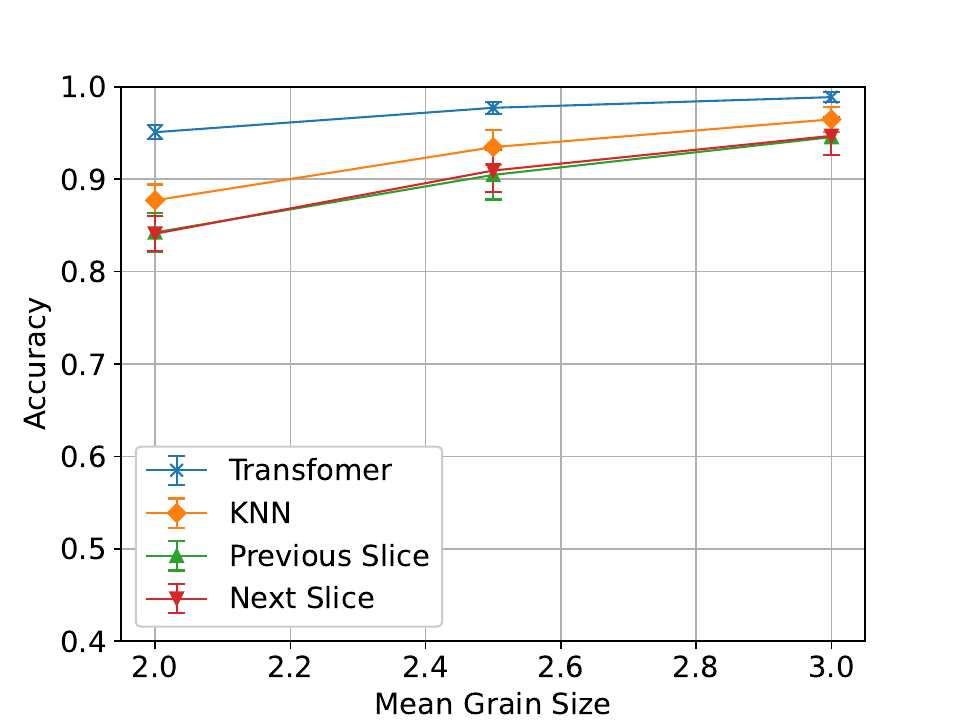}\\
         \caption{Overall Accuracy}
        \label{fig:size_overall_acc_avg}
     \end{subfigure}
     \begin{subfigure}[b]{0.49\linewidth}
         \centering
         \includegraphics[width=1\linewidth]{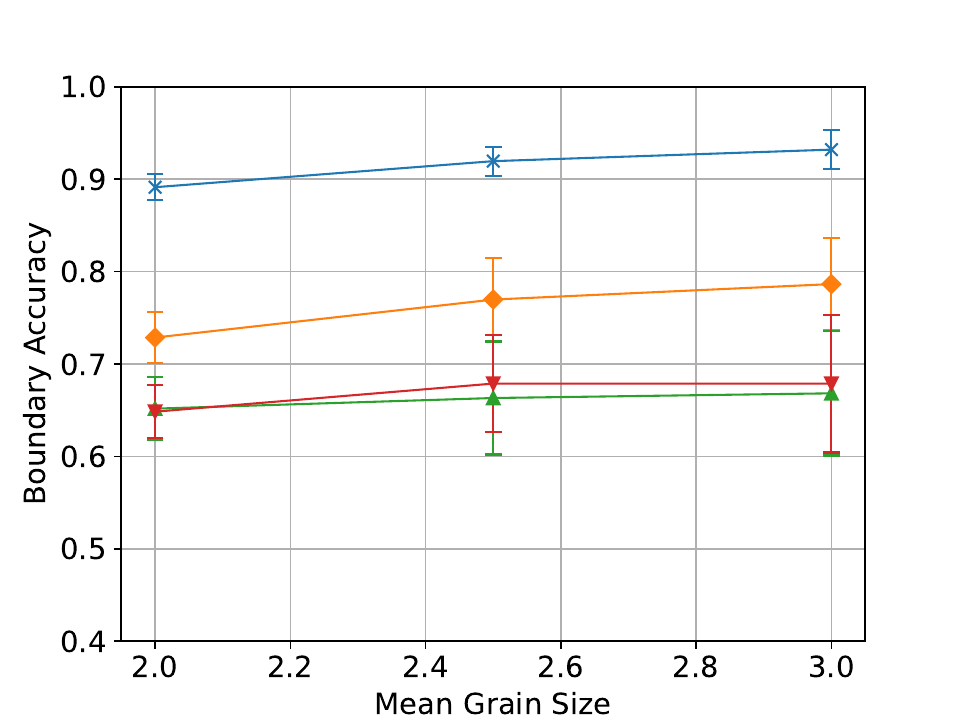} \\
         \caption{Boundary Accuracy}
        \label{fig:size_edge_acc_avg}
     \end{subfigure}
     \begin{subfigure}[b]{0.49\linewidth}
         \centering
         \includegraphics[width=1\linewidth]{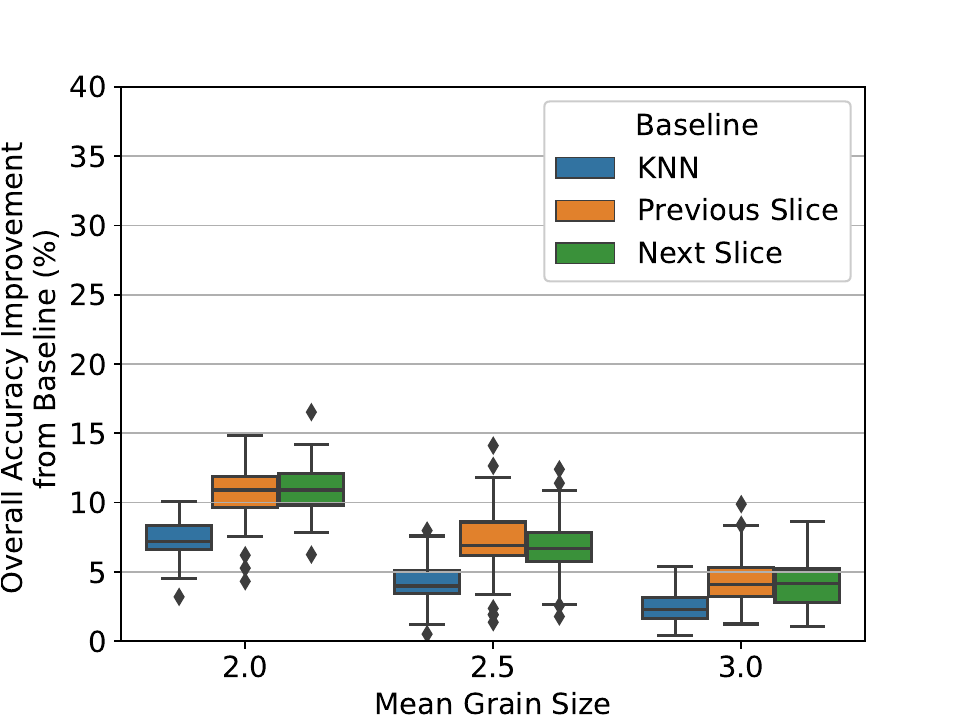}\\
         \caption{Overall Accuracy Improvements}
        \label{fig:size_overall_acc}
     \end{subfigure}
     \begin{subfigure}[b]{0.49\linewidth}
         \centering
         \includegraphics[width=1\linewidth]{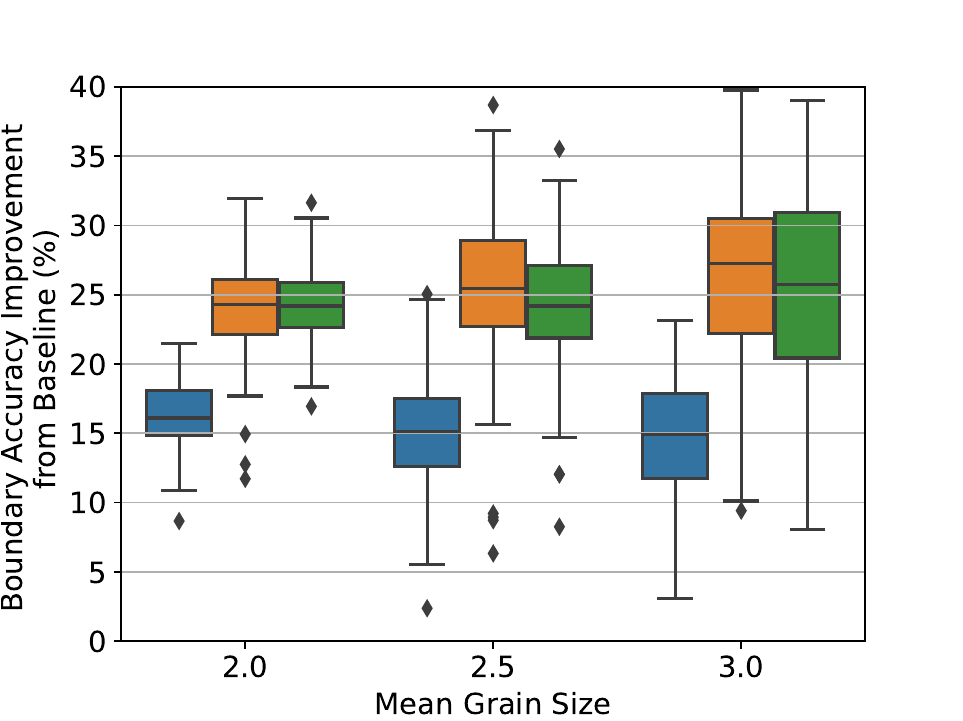} \\
         \caption{Boundary Accuracy Improvements}
        \label{fig:size_edge_acc}
     \end{subfigure}
     \vspace{-0.1in}
     \end{minipage}
        \caption{Overall accuracy (Figure \ref{fig:size_overall_acc_avg}), boundary accuracy (Figure \ref{fig:size_edge_acc_avg}), overall accuracy improvements (Figure \ref{fig:size_overall_acc}), and boundary accuracy improvements (Figure \ref{fig:size_edge_acc}) of synthetic volumes with varying mean grain sizes. Error bars in Figures \ref{fig:size_overall_acc_avg} and \ref{fig:size_edge_acc_avg} represent one standard deviation.}
        \label{fig:size_acc}
\end{figure}

First, we generate 4 independent synthetic volumes of shape $(64, 192, 64)$  for each setting. Each volume is divided into 27 nonoverlapping segments of shape $(64, 7, 64)$ each with one of the five interior slices masked. This results in 108 validation samples for each setting. These validation metrics are displayed in Figure \ref{fig:twins_acc} when varying the mean twins frequency and in Figure \ref{fig:size_acc} when varying mean grain size.

We observe that our method more accurately recovers missing slices than all other baselines for every synthetic validation sample since each slice observed a positive improvement in overall and boundary accuracy. The performance gain is much more apparent for boundary voxels. Furthermore, our transformer performance is much lower variance. Among the baselines, KNN achieves the closest accuracy to our method, but it still underperforms in comparison. As expected, the difference between our method and the baselines diminishes as the scenario get simpler (larger grain sizes and fewer twins) since the baselines are already producing very accurate results.

\subsubsection*{Real EBSD Datasets}
\label{real_experiments}

\begin{figure}
    \begin{minipage}[b]{1\linewidth}
     \centering
     \begin{subfigure}[b]{0.49\linewidth}
         \centering
         \includegraphics[width=1\linewidth]{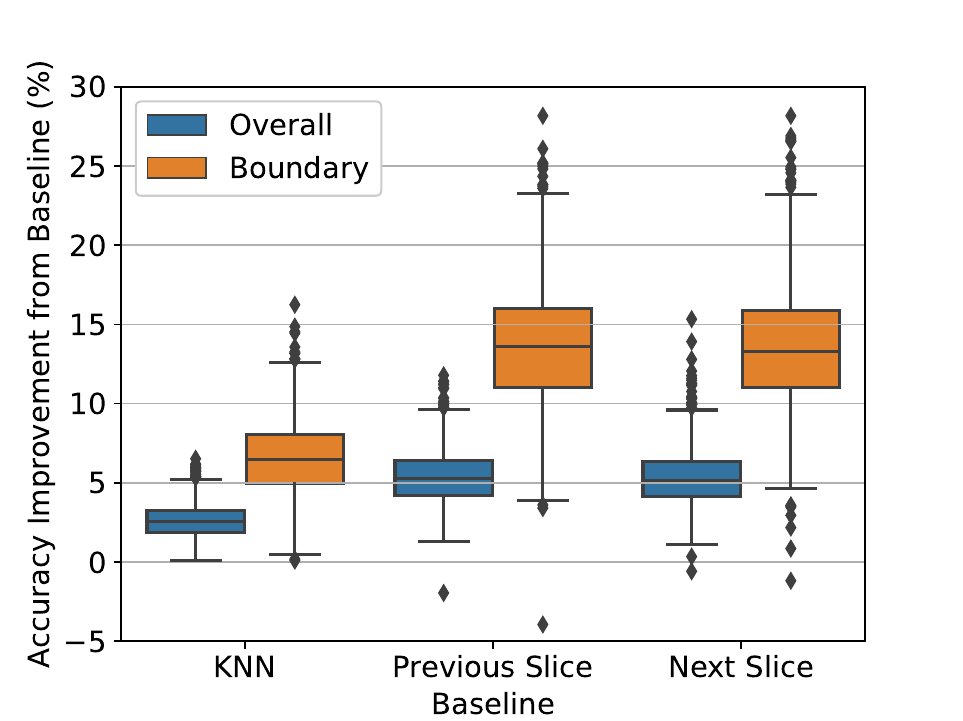}\\
         \caption{IN625}
        \label{fig:in625}
     \end{subfigure}
     \begin{subfigure}[b]{0.49\linewidth}
         \centering
         \includegraphics[width=1\linewidth]{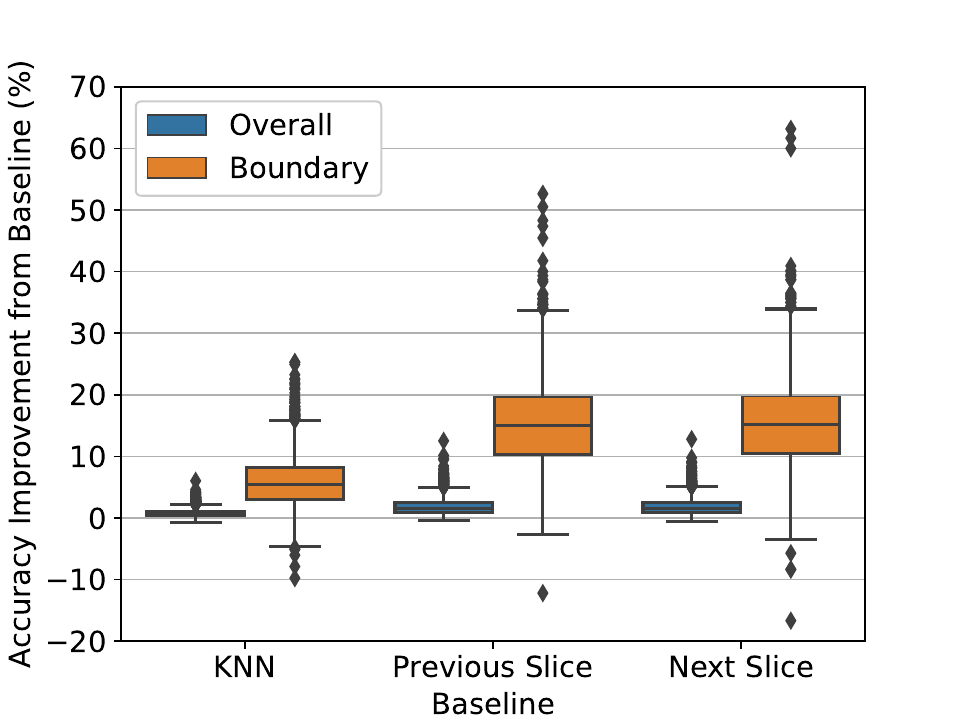} \\
         \caption{IN718}
        \label{fig:in718}
     \end{subfigure}
     \vspace{-0.1in}
     \end{minipage}
        \caption{Overall and boundary accuracy improvements of IN625 and IN718 test sets. Note that the y-axes are on different scales.}
        \label{fig:real}
\end{figure}

\begin{figure}
    \begin{minipage}[b]{1\linewidth}
     \centering
     \begin{subfigure}[b]{0.14\linewidth}
         \centering     
         \includegraphics[width=1\linewidth]{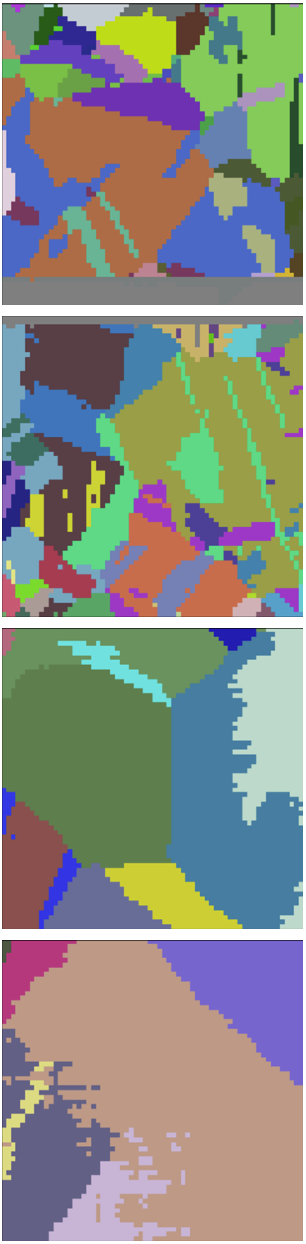}\\
         \caption{}
        \label{fig:output_target}
     \end{subfigure}
    \begin{subfigure}[b]{0.14\linewidth}
         \centering     
         \includegraphics[width=1\linewidth]{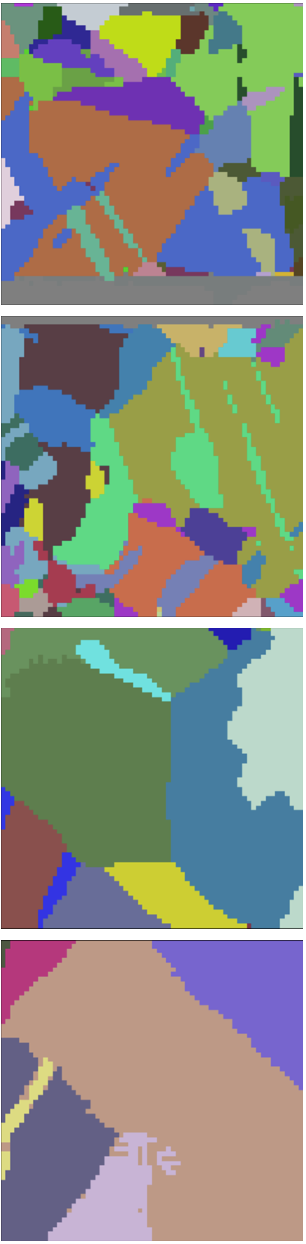}\\
         \caption{}
        \label{fig:output_transformer}
     \end{subfigure}
     \begin{subfigure}[b]{0.14\linewidth}
         \centering     
         \includegraphics[width=1\linewidth]{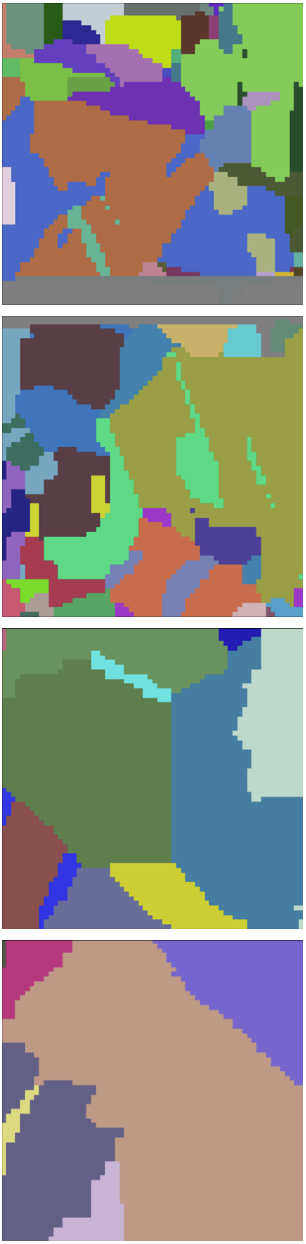}\\
         \caption{}
        \label{fig:output_knn}
     \end{subfigure}
    \begin{subfigure}[b]{0.14\linewidth}
         \centering     
         \includegraphics[width=1\linewidth]{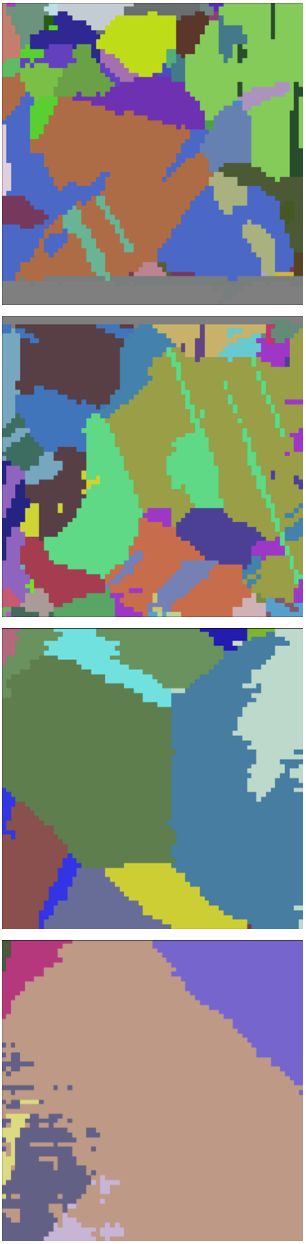}\\
         \caption{}
        \label{fig:output_prev}
     \end{subfigure}
    \begin{subfigure}[b]{0.14\linewidth}
         \centering     
         \includegraphics[width=1\linewidth]{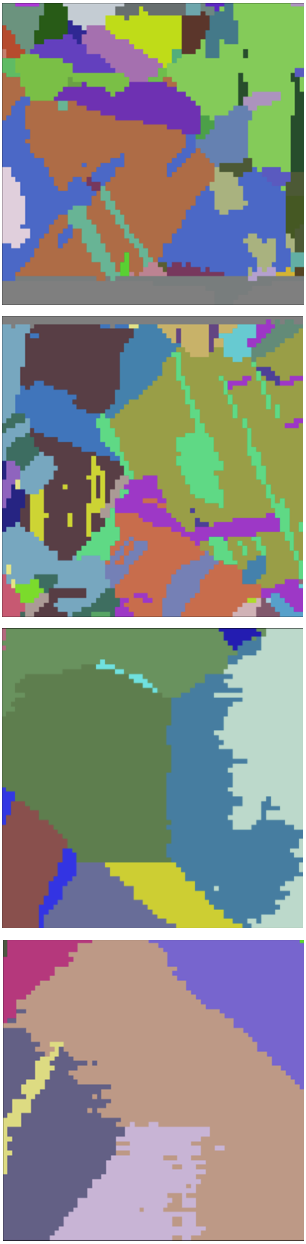}\\
         \caption{}
        \label{fig:output_next}
     \end{subfigure}
     \end{minipage}
        \caption{Four random example predictions of missing slices from IN625 (top two rows) and IN718 (bottom two rows) test sets . Each row is a separate example. Figure \ref{fig:output_target} contains the target; Figures  \ref{fig:output_transformer}, \ref{fig:output_knn}, \ref{fig:output_prev}, and \ref{fig:output_next} are predictions made by our transformer, KNN, the previous slice, and the following slice, respectively.}
        \label{fig:output_examples}
\end{figure}

Now, we seek to understand how well our model transfers to real data by running our trained model on IN625 \cite{menasche2021afrl, shade2019afrl} and IN718 \cite{stinville2022multi, stinville2022multidataset}. For both volumes, we subdivide each volume into nonoverlapping subvolumes such that all slices are of height and width 64 voxels. Then, each subvolume is then further partitioned into nonoverlapping segments of shape $(64, 7, 64, *)$, each representing a single test sample.  Again, each sample contains one masked slice among the five central slices. In the end, there are 800 samples for IN625 and 3298 samples for IN718 after discarding a small set of samples whose missing slice did not have any boundary voxels (these samples would trivially result in exact recovery regardless of the model we use). Average accuracies and accuracy improvements for both volumes are shown in Table \ref{real_acc_table} and Figure \ref{fig:real}, respectively. For recovery examples, see Figure \ref{fig:output_examples}.

\begin{table}[t]
\caption{Average overall accuracy and boundary accuracy across samples of real world datasets, including their standard deviations.}
\label{real_acc_table}
% \vskip 0.15in
\begin{center}
\begin{small}
\begin{sc}
\begin{tabular}{lcccc}
\toprule
Dataset & Model & Avg. Acc. (\%) & Avg. Boundary Acc. (\%) \\
\midrule
\multirow{4}{*}{IN625} &Transformer & \textbf{91.83 $\pm$ 1.98} & \textbf{79.49 $\pm$ 3.20}\\
&KNN & 89.18 $\pm$ 2.48 & 72.86 $\pm$ 3.77\\
&Previous Slice & 86.44 $\pm$ 2.94 & 65.83 $\pm$ 4.73\\
&Next Slice & 86.45 $\pm$ 3.15 & 65.91 $\pm$ 4.98\\
\midrule
\multirow{4}{*}{IN718} &Transformer &\textbf{98.27 $\pm$ 1.03} & \textbf{85.65 $\pm$ 3.49} \\
&KNN & 97.54 $\pm$ 1.52 &79.81 $\pm$ 5.49 \\
&Previous Slice &96.40 $\pm$ 2.23 &70.43 $\pm$ 8.50 \\
&Next Slice &96.39 $\pm$ 2.22 &70.25 $\pm$ 8.47 \\
\bottomrule
\end{tabular}
\end{sc}
\end{small}
\end{center}
\vskip -0.1in
\end{table}

Surprisingly, even though our model is trained exclusively on synthetic data, we observe it transfers well to real EBSD data, as it still outperforms every baseline on nearly every sample, and the interquartile range is positive. Again, the difference is accentuated for boundary voxels. Test results for the IN718 dataset had much higher variance, likely due to each slice having proportionally fewer boundary voxels than IN625 slices. Qualitatively, our model has a stronger capability to recovery thin features than KNN, which visually tends to ignore more subtle structures.

% \begin{figure}
%     \begin{minipage}[b]{1\linewidth}
%      \centering
%      \begin{subfigure}[b]{0.47\linewidth}
%          \centering
%          \includegraphics[width=8cm]{figures/overall_midas_reverse.pdf}\\
%          \caption{Overall Accuracy}
%         \label{fig:midas_overall_acc}
%      \end{subfigure}
%      \begin{subfigure}[b]{0.47\linewidth}
%          \centering
%          \includegraphics[width=8cm]{figures/edge_midas_reverse.pdf} \\
%          \caption{Accuracy Along Boundary Voxels}
%         \label{fig:midas_edge_acc}
%      \end{subfigure}
%      \end{minipage}
%         \caption{Box plot of the accuracies of the recovered slice overall (left) and conditioned on edge voxels (right).}
%         \label{fig:midas_acc}
% \end{figure}

% Even though our model is trained exclusively on synthetic data, we observe it transfers well to the MIDAS dataset, as it still outperforms all baselines. We note that our model is able to preserve very thin features in the missing slices, whereas KNN smooths them out.

\subsection*{Limitations}

% Due to the presence of heavy left tails in Figure \ref{fig:midas_acc}, we seek to understand what makes these missing slices difficult to recover. 
While our method provides superior results compared to the baselines, we also seek to understand the circumstance in which it  may underperform. One challenge is rapid changes between slices which make up a small minority of test inputs. For instance, if the $k$-th slice is missing and the set of grains present in the $(k-1)$-th is different from the set of grains in $(k+1)$-th slice, the model has a lot of freedom to decide on which ones are present and to what degree in the missing slice. A case in which this can arise is when the faces of grains or twins are perpendicular to the slicing direction. However, we observe this is an issue for all methods. Using the second row of Figure \ref{fig:real} as an example, the bright green grain is a large crescent shape on the slice before the missing slice but is hardly present in the following slice. Our model and KNN strives to find some middleground, but ultimately, both misclassifies many of the voxels. A similar scene plays out in the fourth row of Figure \ref{fig:real}. Thus, future work includes designing a better loss function to mitigate errors for this or emphasizing these scenarios during training. 

% \begin{figure}
% \centering
%      \includegraphics[width=8cm]{figures/rapid/rapid_midas.png}\\
%      \caption{Example of rapid changes between slices. The top and bottom images are observed preceding and succeeding slices, respectively. The left center slice is masked. The center and right images are predictions made by our model and KNN, respectively.}
%     \label{fig:rapid_midas}
% \end{figure}

\begin{figure}
\centering
\includegraphics[width= 0.85 \linewidth]{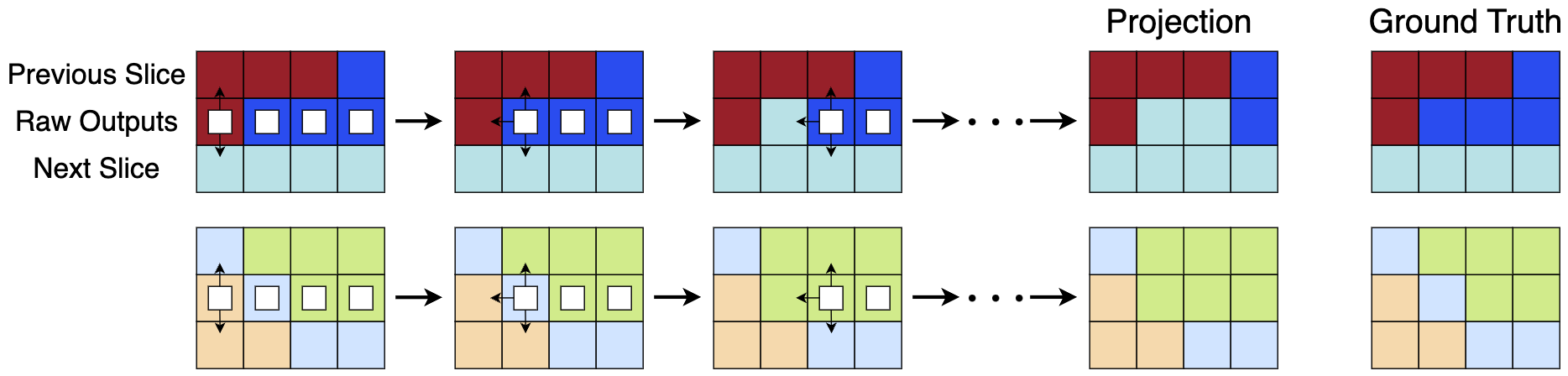}\\
\caption{Matrix examples where projection to observed and previously projected neighbors is suboptimal, even when giving it the ground truth. Projection involves using the neighboring slices to predict the center slice. White squares indicate which pixels have not been projected yet, and the black arrows point to the possible neighbors that the pixel will choose to project to. The top row shows an example where the projection operation smooths out very thin features that lie completely in the missing slice. The bottom row shows an example where the projection operation may disconnect a grain.}
\label{fig:projection_smoothing}
\end{figure}

Another limitation arises from the locality of our projection method. Recall our local projection method projects a voxel to have the ID as one of its observed or previously assigned voxels in order to encourage grain connectedness. This means long range dependencies may be ignored in favor of more smooth structures. Furthermore, connectedness is not guaranteed for very thin features at an angle. Examples of both edge cases for matrices are illustrated in Figure \ref{fig:projection_smoothing}. Though these scenarios are fairly rare in practice, further work could be done to improve our projection algorithm. 

%% file: conclusion.tex
\section*{Conclusion}
\label{sec:conclusion}

We have presented a novel method using transformers followed by a projection algorithm to recover missing 3D EBSD data which vastly outperforms all baselines. Notably, even though our model is trained on synthetic data, it still recovers more accurate slices for real 3D EBSD data than the baselines by a wide margin, making it a powerful data processing tool for faulty 3D EBSD readings. Furthermore, our model opens the possibility to skip every fourth slice during data collection (by using every skipped slice and the three collected slices on each side as the input to our model), potentially reducing the collection time by 25\%. Future work involves addressing the limitations, scaling our method, and considering more general cases, such as altering the projection algorithm to apply to consecutive missing slices within a sample. In terms of scaling, we hope to observe emergent behavior similar to scaling LLMs and their training data, which have seen vast improvements to performance and versatility \cite{wei2022emergent}. Beyond EBSD, it would be interesting to investigate our method's applicability for other high dimensional material science datasets.

%% file: main.bbl
\begin{thebibliography}{10}
\urlstyle{rm}
\expandafter\ifx\csname url\endcsname\relax
  \def\url#1{\texttt{#1}}\fi
\expandafter\ifx\csname urlprefix\endcsname\relax\def\urlprefix{URL }\fi
\expandafter\ifx\csname doiprefix\endcsname\relax\def\doiprefix{DOI: }\fi
\providecommand{\bibinfo}[2]{#2}
\providecommand{\eprint}[2][]{\url{#2}}

\bibitem{uchic_automated_2016}
\bibinfo{author}{Uchic, M.} \emph{et~al.}
\newblock \bibinfo{title}{An {Automated} {Multi}-{Modal} {Serial} {Sectioning}
  {System} for {Characterization} of {Grain}-{Scale} {Microstructures} in
  {Engineering} {Materials}}.
\newblock In \bibinfo{editor}{De~Graef, M.}, \bibinfo{editor}{Poulsen, H.~F.},
  \bibinfo{editor}{Lewis, A.}, \bibinfo{editor}{Simmons, J.} \&
  \bibinfo{editor}{Spanos, G.} (eds.) \emph{\bibinfo{booktitle}{Proceedings of
  the 1st {International} {Conference} on {3D} {Materials} {Science}}},
  \bibinfo{pages}{195--202}, \doiprefix\url{10.1007/978-3-319-48762-5_30}
  (\bibinfo{publisher}{Springer International Publishing},
  \bibinfo{address}{Cham}, \bibinfo{year}{2016}).

\bibitem{chapman2021afrl}
\bibinfo{author}{Chapman, M.~G.} \emph{et~al.}
\newblock \bibinfo{journal}{\bibinfo{title}{Afrl additive manufacturing
  modeling series: challenge 4, 3d reconstruction of an in625 high-energy
  diffraction microscopy sample using multi-modal serial sectioning}}.
\newblock {\emph{\JournalTitle{Integrating Materials and Manufacturing
  Innovation}}} \textbf{\bibinfo{volume}{10}}, \bibinfo{pages}{129--141}
  (\bibinfo{year}{2021}).

\bibitem{polonsky_scan_2022}
\bibinfo{author}{Polonsky, A.~T.} \emph{et~al.}
\newblock \bibinfo{journal}{\bibinfo{title}{Scan strategies in {EBM}-printed
  {IN718} and the physics of bulk {3D} microstructure development}}.
\newblock {\emph{\JournalTitle{Materials Characterization}}}
  \textbf{\bibinfo{volume}{190}}, \bibinfo{pages}{112043},
  \doiprefix\url{10.1016/j.matchar.2022.112043} (\bibinfo{year}{2022}).

\bibitem{polonsky_three-dimensional_2019}
\bibinfo{author}{Polonsky, A.~T.} \emph{et~al.}
\newblock \bibinfo{journal}{\bibinfo{title}{Three-dimensional {Analysis} and
  {Reconstruction} of {Additively} {Manufactured} {Materials} in the
  {Cloud}-{Based} {BisQue} {Infrastructure}}}.
\newblock {\emph{\JournalTitle{Integrating Materials and Manufacturing
  Innovation}}} \textbf{\bibinfo{volume}{8}}, \bibinfo{pages}{37--51},
  \doiprefix\url{10.1007/s40192-019-00126-7} (\bibinfo{year}{2019}).

\bibitem{jolley_application_2021}
\bibinfo{author}{Jolley, B.~R.}, \bibinfo{author}{Uchic, M.~D.},
  \bibinfo{author}{Sparkman, D.}, \bibinfo{author}{Chapman, M.} \&
  \bibinfo{author}{Schwalbach, E.~J.}
\newblock \bibinfo{journal}{\bibinfo{title}{Application of {Serial}
  {Sectioning} to {Evaluate} the {Performance} of x-ray {Computed} {Tomography}
  for {Quantitative} {Porosity} {Measurements} in {Additively} {Manufactured}
  {Metals}}}.
\newblock {\emph{\JournalTitle{JOM}}}
  \doiprefix\url{10.1007/s11837-021-04863-z} (\bibinfo{year}{2021}).

\bibitem{nguyen_alignment_2021}
\bibinfo{author}{Nguyen, L.~T.} \& \bibinfo{author}{Rowenhorst, D.~J.}
\newblock \bibinfo{journal}{\bibinfo{title}{The {Alignment} and {Fusion} of
  {Multimodal} {3D} {Serial} {Sectioning} {Datasets}}}.
\newblock {\emph{\JournalTitle{JOM}}} \textbf{\bibinfo{volume}{73}},
  \bibinfo{pages}{3272--3284}, \doiprefix\url{10.1007/s11837-021-04865-x}
  (\bibinfo{year}{2021}).

\bibitem{kotula2006tomographic}
\bibinfo{author}{Kotula, P.~G.}, \bibinfo{author}{Keenan, M.~R.} \&
  \bibinfo{author}{Michael, J.~R.}
\newblock \bibinfo{journal}{\bibinfo{title}{Tomographic spectral imaging with
  multivariate statistical analysis: comprehensive 3d microanalysis}}.
\newblock {\emph{\JournalTitle{Microscopy and Microanalysis}}}
  \textbf{\bibinfo{volume}{12}}, \bibinfo{pages}{36--48}
  (\bibinfo{year}{2006}).

\bibitem{calcagnotto2010orientation}
\bibinfo{author}{Calcagnotto, M.}, \bibinfo{author}{Ponge, D.},
  \bibinfo{author}{Demir, E.} \& \bibinfo{author}{Raabe, D.}
\newblock \bibinfo{journal}{\bibinfo{title}{Orientation gradients and
  geometrically necessary dislocations in ultrafine grained dual-phase steels
  studied by 2d and 3d ebsd}}.
\newblock {\emph{\JournalTitle{Materials Science and Engineering: A}}}
  \textbf{\bibinfo{volume}{527}}, \bibinfo{pages}{2738--2746}
  (\bibinfo{year}{2010}).

\bibitem{naragani2017investigation}
\bibinfo{author}{Naragani, D.} \emph{et~al.}
\newblock \bibinfo{journal}{\bibinfo{title}{Investigation of fatigue crack
  initiation from a non-metallic inclusion via high energy x-ray diffraction
  microscopy}}.
\newblock {\emph{\JournalTitle{Acta Materialia}}}
  \textbf{\bibinfo{volume}{137}}, \bibinfo{pages}{71--84}
  (\bibinfo{year}{2017}).

\bibitem{sandgren2016characterization}
\bibinfo{author}{Sandgren, H.~R.} \emph{et~al.}
\newblock \bibinfo{journal}{\bibinfo{title}{Characterization of fatigue crack
  growth behavior in lens fabricated ti-6al-4v using high-energy synchrotron
  x-ray microtomography}}.
\newblock {\emph{\JournalTitle{Additive Manufacturing}}}
  \textbf{\bibinfo{volume}{12}}, \bibinfo{pages}{132--141}
  (\bibinfo{year}{2016}).

\bibitem{wilson2006three}
\bibinfo{author}{Wilson, J.~R.} \emph{et~al.}
\newblock \bibinfo{journal}{\bibinfo{title}{Three-dimensional reconstruction of
  a solid-oxide fuel-cell anode}}.
\newblock {\emph{\JournalTitle{Nature materials}}}
  \textbf{\bibinfo{volume}{5}}, \bibinfo{pages}{541--544}
  (\bibinfo{year}{2006}).

\bibitem{teferra_optimizing_2021}
\bibinfo{author}{Teferra, K.} \& \bibinfo{author}{Rowenhorst, D.~J.}
\newblock \bibinfo{journal}{\bibinfo{title}{Optimizing the cellular automata
  finite element model for additive manufacturing to simulate large
  microstructures}}.
\newblock {\emph{\JournalTitle{Acta Materialia}}} \bibinfo{pages}{116930},
  \doiprefix\url{10.1016/j.actamat.2021.116930} (\bibinfo{year}{2021}).

\bibitem{vaswani2017attention}
\bibinfo{author}{Vaswani, A.} \emph{et~al.}
\newblock \bibinfo{journal}{\bibinfo{title}{Attention is all you need}}.
\newblock {\emph{\JournalTitle{Advances in neural information processing
  systems}}} \textbf{\bibinfo{volume}{30}} (\bibinfo{year}{2017}).

\bibitem{khan2022transformers}
\bibinfo{author}{Khan, S.} \emph{et~al.}
\newblock \bibinfo{journal}{\bibinfo{title}{Transformers in vision: A survey}}.
\newblock {\emph{\JournalTitle{ACM computing surveys (CSUR)}}}
  \textbf{\bibinfo{volume}{54}}, \bibinfo{pages}{1--41} (\bibinfo{year}{2022}).

\bibitem{latif2023transformers}
\bibinfo{author}{Latif, S.} \emph{et~al.}
\newblock \bibinfo{journal}{\bibinfo{title}{Transformers in speech processing:
  A survey}}.
\newblock {\emph{\JournalTitle{arXiv preprint arXiv:2303.11607}}}
  (\bibinfo{year}{2023}).

\bibitem{zhang2023applications}
\bibinfo{author}{Zhang, S.} \emph{et~al.}
\newblock \bibinfo{journal}{\bibinfo{title}{Applications of transformer-based
  language models in bioinformatics: A survey}}.
\newblock {\emph{\JournalTitle{Bioinformatics Advances}}}
  (\bibinfo{year}{2023}).

\bibitem{devlin2018bert}
\bibinfo{author}{Devlin, J.}, \bibinfo{author}{Chang, M.-W.},
  \bibinfo{author}{Lee, K.} \& \bibinfo{author}{Toutanova, K.}
\newblock \bibinfo{journal}{\bibinfo{title}{Bert: Pre-training of deep
  bidirectional transformers for language understanding}}.
\newblock {\emph{\JournalTitle{arXiv preprint arXiv:1810.04805}}}
  (\bibinfo{year}{2018}).

\bibitem{radford2018improving}
\bibinfo{author}{Radford, A.}, \bibinfo{author}{Narasimhan, K.},
  \bibinfo{author}{Salimans, T.}, \bibinfo{author}{Sutskever, I.} \emph{et~al.}
\newblock \bibinfo{title}{Improving language understanding by generative
  pre-training} (\bibinfo{year}{2018}).

\bibitem{kalyan2021ammus}
\bibinfo{author}{Kalyan, K.~S.}, \bibinfo{author}{Rajasekharan, A.} \&
  \bibinfo{author}{Sangeetha, S.}
\newblock \bibinfo{journal}{\bibinfo{title}{Ammus: A survey of
  transformer-based pretrained models in natural language processing}}.
\newblock {\emph{\JournalTitle{arXiv preprint arXiv:2108.05542}}}
  (\bibinfo{year}{2021}).

\bibitem{zhang2022opt}
\bibinfo{author}{Zhang, S.} \emph{et~al.}
\newblock \bibinfo{journal}{\bibinfo{title}{Opt: Open pre-trained transformer
  language models}}.
\newblock {\emph{\JournalTitle{arXiv preprint arXiv:2205.01068}}}
  (\bibinfo{year}{2022}).

\bibitem{brown2020language}
\bibinfo{author}{Brown, T.} \emph{et~al.}
\newblock \bibinfo{journal}{\bibinfo{title}{Language models are few-shot
  learners}}.
\newblock {\emph{\JournalTitle{Advances in neural information processing
  systems}}} \textbf{\bibinfo{volume}{33}}, \bibinfo{pages}{1877--1901}
  (\bibinfo{year}{2020}).

\bibitem{touvron2023llama}
\bibinfo{author}{Touvron, H.} \emph{et~al.}
\newblock \bibinfo{title}{Llama: Open and efficient foundation language models}
  (\bibinfo{year}{2023}).
\newblock \eprint{2302.13971}.

\bibitem{pathak2016context}
\bibinfo{author}{Pathak, D.}, \bibinfo{author}{Krahenbuhl, P.},
  \bibinfo{author}{Donahue, J.}, \bibinfo{author}{Darrell, T.} \&
  \bibinfo{author}{Efros, A.~A.}
\newblock \bibinfo{title}{Context encoders: Feature learning by inpainting}.
\newblock In \emph{\bibinfo{booktitle}{Proceedings of the IEEE conference on
  computer vision and pattern recognition}}, \bibinfo{pages}{2536--2544}
  (\bibinfo{year}{2016}).

\bibitem{he2022masked}
\bibinfo{author}{He, K.} \emph{et~al.}
\newblock \bibinfo{title}{Masked autoencoders are scalable vision learners}.
\newblock In \emph{\bibinfo{booktitle}{Proceedings of the IEEE/CVF Conference
  on Computer Vision and Pattern Recognition}}, \bibinfo{pages}{16000--16009}
  (\bibinfo{year}{2022}).

\bibitem{kong2023understanding}
\bibinfo{author}{Kong, L.} \emph{et~al.}
\newblock \bibinfo{title}{Understanding masked autoencoders via hierarchical
  latent variable models}.
\newblock In \emph{\bibinfo{booktitle}{Proceedings of the IEEE/CVF Conference
  on Computer Vision and Pattern Recognition}}, \bibinfo{pages}{7918--7928}
  (\bibinfo{year}{2023}).

\bibitem{chang2019free}
\bibinfo{author}{Chang, Y.-L.}, \bibinfo{author}{Liu, Z.~Y.},
  \bibinfo{author}{Lee, K.-Y.} \& \bibinfo{author}{Hsu, W.}
\newblock \bibinfo{title}{Free-form video inpainting with 3d gated convolution
  and temporal patchgan}.
\newblock In \emph{\bibinfo{booktitle}{Proceedings of the IEEE/CVF
  International Conference on Computer Vision}}, \bibinfo{pages}{9066--9075}
  (\bibinfo{year}{2019}).

\bibitem{liu2021decoupled}
\bibinfo{author}{Liu, R.} \emph{et~al.}
\newblock \bibinfo{journal}{\bibinfo{title}{Decoupled spatial-temporal
  transformer for video inpainting}}.
\newblock {\emph{\JournalTitle{arXiv preprint arXiv:2104.06637}}}
  (\bibinfo{year}{2021}).

\bibitem{dosovitskiy2020image}
\bibinfo{author}{Dosovitskiy, A.} \emph{et~al.}
\newblock \bibinfo{journal}{\bibinfo{title}{An image is worth 16x16 words:
  Transformers for image recognition at scale}}.
\newblock {\emph{\JournalTitle{arXiv preprint arXiv:2010.11929}}}
  (\bibinfo{year}{2020}).

\bibitem{dong2023deep}
\bibinfo{author}{Dong, H.}, \bibinfo{author}{Shah, M.},
  \bibinfo{author}{Donegan, S.} \& \bibinfo{author}{Chi, Y.}
\newblock \bibinfo{title}{Deep unfolded tensor robust pca with self-supervised
  learning}.
\newblock In \emph{\bibinfo{booktitle}{ICASSP 2023-2023 IEEE International
  Conference on Acoustics, Speech and Signal Processing (ICASSP)}},
  \bibinfo{pages}{1--5} (\bibinfo{organization}{IEEE}, \bibinfo{year}{2023}).

\bibitem{wang2020linformer}
\bibinfo{author}{Wang, S.}, \bibinfo{author}{Li, B.~Z.},
  \bibinfo{author}{Khabsa, M.}, \bibinfo{author}{Fang, H.} \&
  \bibinfo{author}{Ma, H.}
\newblock \bibinfo{journal}{\bibinfo{title}{Linformer: Self-attention with
  linear complexity}}.
\newblock {\emph{\JournalTitle{arXiv preprint arXiv:2006.04768}}}
  (\bibinfo{year}{2020}).

\bibitem{kitaev2020reformer}
\bibinfo{author}{Kitaev, N.}, \bibinfo{author}{Kaiser, {\L}.} \&
  \bibinfo{author}{Levskaya, A.}
\newblock \bibinfo{journal}{\bibinfo{title}{Reformer: The efficient
  transformer}}.
\newblock {\emph{\JournalTitle{arXiv preprint arXiv:2001.04451}}}
  (\bibinfo{year}{2020}).

\bibitem{zaheer2020big}
\bibinfo{author}{Zaheer, M.} \emph{et~al.}
\newblock \bibinfo{journal}{\bibinfo{title}{Big bird: Transformers for longer
  sequences}}.
\newblock {\emph{\JournalTitle{Advances in Neural Information Processing
  Systems}}} \textbf{\bibinfo{volume}{33}}, \bibinfo{pages}{17283--17297}
  (\bibinfo{year}{2020}).

\bibitem{tay2022efficient}
\bibinfo{author}{Tay, Y.}, \bibinfo{author}{Dehghani, M.},
  \bibinfo{author}{Bahri, D.} \& \bibinfo{author}{Metzler, D.}
\newblock \bibinfo{journal}{\bibinfo{title}{Efficient transformers: A survey}}.
\newblock {\emph{\JournalTitle{ACM Computing Surveys}}}
  \textbf{\bibinfo{volume}{55}}, \bibinfo{pages}{1--28} (\bibinfo{year}{2022}).

\bibitem{ho2019axial}
\bibinfo{author}{Ho, J.}, \bibinfo{author}{Kalchbrenner, N.},
  \bibinfo{author}{Weissenborn, D.} \& \bibinfo{author}{Salimans, T.}
\newblock \bibinfo{journal}{\bibinfo{title}{Axial attention in multidimensional
  transformers}}.
\newblock {\emph{\JournalTitle{arXiv preprint arXiv:1912.12180}}}
  (\bibinfo{year}{2019}).

\bibitem{wang2020axial}
\bibinfo{author}{Wang, H.} \emph{et~al.}
\newblock \bibinfo{title}{Axial-deeplab: Stand-alone axial-attention for
  panoptic segmentation}.
\newblock In \emph{\bibinfo{booktitle}{Computer Vision--ECCV 2020: 16th
  European Conference, Glasgow, UK, August 23--28, 2020, Proceedings, Part
  IV}}, \bibinfo{pages}{108--126} (\bibinfo{organization}{Springer},
  \bibinfo{year}{2020}).

\bibitem{menasche2021afrl}
\bibinfo{author}{Menasche, D.~B.} \emph{et~al.}
\newblock \bibinfo{journal}{\bibinfo{title}{Afrl additive manufacturing
  modeling series: challenge 4, in situ mechanical test of an in625 sample with
  concurrent high-energy diffraction microscopy characterization}}.
\newblock {\emph{\JournalTitle{Integrating Materials and Manufacturing
  Innovation}}} \textbf{\bibinfo{volume}{10}}, \bibinfo{pages}{338--347}
  (\bibinfo{year}{2021}).

\bibitem{shade2019afrl}
\bibinfo{author}{Shade, P.~A.} \emph{et~al.}
\newblock \bibinfo{title}{Afrl am modeling challenge series: Challenge 4 data
  package}, \doiprefix\url{10.18126/K5R2-32IU} (\bibinfo{year}{2019}).

\bibitem{stinville2022multi}
\bibinfo{author}{Stinville, J.} \emph{et~al.}
\newblock \bibinfo{journal}{\bibinfo{title}{Multi-modal dataset of a
  polycrystalline metallic material: 3d microstructure and deformation
  fields}}.
\newblock {\emph{\JournalTitle{Scientific Data}}} \textbf{\bibinfo{volume}{9}},
  \bibinfo{pages}{460} (\bibinfo{year}{2022}).

\bibitem{stinville2022multidataset}
\bibinfo{author}{Stinville, J.} \emph{et~al.}
\newblock \bibinfo{title}{Multi-modal dataset of a polycrystalline metallic
  material: 3d microstructure and deformation fields},
  \doiprefix\url{10.5061/dryad.83bk3j9sj} (\bibinfo{year}{2022}).

\bibitem{rocsca2014new}
\bibinfo{author}{Ro{\c{s}}ca, D.}, \bibinfo{author}{Morawiec, A.} \&
  \bibinfo{author}{De~Graef, M.}
\newblock \bibinfo{journal}{\bibinfo{title}{A new method of constructing a grid
  in the space of 3d rotations and its applications to texture analysis}}.
\newblock {\emph{\JournalTitle{Modelling and Simulation in Materials Science
  and Engineering}}} \textbf{\bibinfo{volume}{22}}, \bibinfo{pages}{075013}
  (\bibinfo{year}{2014}).

\bibitem{groeber2014dream}
\bibinfo{author}{Groeber, M.~A.} \& \bibinfo{author}{Jackson, M.~A.}
\newblock \bibinfo{journal}{\bibinfo{title}{Dream. 3d: a digital representation
  environment for the analysis of microstructure in 3d}}.
\newblock {\emph{\JournalTitle{Integrating materials and manufacturing
  innovation}}} \textbf{\bibinfo{volume}{3}}, \bibinfo{pages}{56--72}
  (\bibinfo{year}{2014}).

\bibitem{donegan2013ev}
\bibinfo{author}{Donegan, S.}, \bibinfo{author}{Tucker, J.},
  \bibinfo{author}{Rollett, A.}, \bibinfo{author}{Barmak, K.} \&
  \bibinfo{author}{M, G.}
\newblock \bibinfo{journal}{\bibinfo{title}{Extreme value analysis of tail
  departure from log-normality in experimental and simulated grain size
  distributions}}.
\newblock {\emph{\JournalTitle{Acta Materialia}}}
  \textbf{\bibinfo{volume}{61}}, \bibinfo{pages}{5595--5604}
  (\bibinfo{year}{2013}).

\bibitem{wei2022emergent}
\bibinfo{author}{Wei, J.} \emph{et~al.}
\newblock \bibinfo{journal}{\bibinfo{title}{Emergent abilities of large
  language models}}.
\newblock {\emph{\JournalTitle{arXiv preprint arXiv:2206.07682}}}
  (\bibinfo{year}{2022}).

\end{thebibliography}
